\documentclass{article}
\usepackage[a4paper, 
            left=2cm, 
            right=2cm, 
            top=3cm, 
            bottom=3cm]{geometry}

\usepackage{silence}
\usepackage{layouts}
\usepackage{utils/algorithm}
\usepackage{utils/algorithmic}
\usepackage{pdfpages}

\usepackage{titlesec} 
\usepackage{setspace} 
\usepackage[strict]{changepage}

\usepackage[round, sort]{natbib}
\usepackage{hyperref} 
\hypersetup{
	final=true,
	plainpages=false,
	pdfstartview=FitV,
	pdftoolbar=true,
	pdfmenubar=true,
	bookmarksopen=true,
	bookmarksnumbered=true,
	breaklinks=true,
	linktocpage,
	colorlinks=true,
	linkcolor=blue,
	urlcolor=ForestGreen,
	citecolor=blue,
	anchorcolor=ForestGreen,
	pdfpagelayout=TwoPageRight,
	pdftitle={Inductive Bias in Machine Learning},
	pdfauthor={Luca Silvester Rendsburg},
}

\usepackage{amsmath, amsfonts, amsthm}
\usepackage{mathtools}

\usepackage{thmtools}
\usepackage{thm-restate}

\usepackage{framed}

\usepackage{enumitem}
\usepackage{graphicx}
\usepackage{subcaption}
\usepackage{booktabs} 
\usepackage{makecell}

\usepackage{tikz}
\usetikzlibrary{cd, bayesnet, arrows}
\usetikzlibrary{decorations.shapes}
\usetikzlibrary{patterns,shapes.arrows}
\usepackage{pgfplots}
\pgfplotsset{compat=newest}

\usepackage[framemethod=tikz]{mdframed}

\makeatletter
\def\th@plain{%
  \thm@notefont{}
  \itshape 
}
\def\th@definition{%
  \thm@notefont{}
  \normalfont 
}
\makeatother

\theoremstyle{definition}

\theoremstyle{plain}

\title{Towards understanding Diffusion Models (on Graphs)}
\author{Solveig Klepper}
\date{November 2023}

\begin{document}

\maketitle

\begin{abstract}
Diffusion models have emerged from various theoretical and methodological perspectives, each offering unique insights into their underlying principles. In this work, we provide an overview of the most prominent approaches, drawing attention to their striking analogies -- namely, how seemingly diverse methodologies converge to a similar mathematical formulation of the core problem.
While our ultimate goal is to understand these models in the context of graphs, we begin by conducting experiments in a simpler setting to build foundational insights. Through an empirical investigation of different diffusion and sampling techniques, we explore three critical questions: (1) What role does noise play in these models? (2) How significantly does the choice of the sampling method affect outcomes? (3) What function is the neural network approximating, and is high complexity necessary for optimal performance? Our findings aim to enhance the understanding of diffusion models and in the long run their application in graph machine learning.

\end{abstract}

\section{Continouos Diffusion Models}

\begin{figure}[ht]
    \centering
    \includegraphics[scale=0.7]{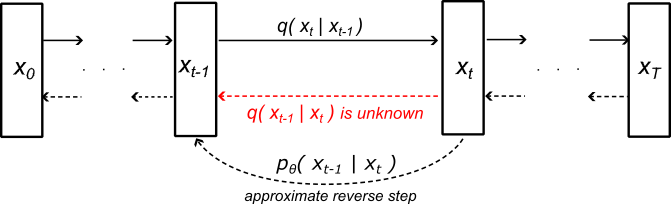}
    \caption{General idea of denoising diffusion models. The forward process is modelled by a Markov process. The reverse process is unknown and needs to be approximated; this is usually done with a neural network.}
    \label{fig:intuition:annotation}
\end{figure}

In physics, diffusion captures the overall movement of particles, such as atoms, from areas of higher concentration to those of lower concentration. Consider the analogy of dropping a small amount of paint into a glass of water. Initially, the paint is concentrated in one location, but over time, it diffuses throughout the water until it reaches a state of equilibrium. The intriguing question arises: Can we reverse this diffusion process? Unfortunately, such a reversal proves impossible in most cases.

Despite the impossibility of reversing diffusion, a field of study known as diffusion models exists. These models aim to capture the dynamics of this diffusion phenomenon and are based on the idea of \textit{approximately} undoing this process. Empirically, they achieve surprisingly good results when sampling new data points with similar properties.

From a practical point of view, diffusion models are generative models that aim to create new samples from an unknown and often complex underlying distribution. Usually, the only information about the target distribution is training data points originating from it. However, directly approximating this training distribution is challenging, so diffusion models systematically decompose the process into incremental steps. Due to the incremental diffusion, the model learns to predict a distribution not only for clean training data but also for a set of distributions generated by gradually adding noise to the training data. This way, the model can learn and improve itself over these steps. This results in high-quality samples. In this context of a chaotic system, each datapoint $x_t$ progressively loses its distinguishable features as the time step $t$ increases. As the number of diffusion steps approaches infinity ($T \to \infty$), the terminal state $x_T$ converges to an isotropic Gaussian distribution, showing the system attained a state of equilibrium.

\subsection{Diffusion Models}

In the past few years, various generative models using the concept of diffusion have been introduced. Different methodologies end up with more or less the same mathematical formulation of the underlying problem.

\subsubsection{Langevin Dynamics}

Inspired by the principles of a molecule diffusing in a liquid, the Langevin formula mathematically captures the diffusion process. The key parameters are the particle mass $m$, the damping coefficient $\lambda$, velocity $v$, and a noise term $\eta$ representing collisions with surrounding molecules.

\begin{equation}\label{eq:langevin}
    m \frac{dv}{dt} = - \lambda v + \eta(t)
\end{equation}

In the context of diffusion models, we describe the forward process similarly. 

\begin{equation}\label{eq:langevinddpm}
    \frac{d \mathbf{x}(t)}{d t} = \mathbf{x}(t) + g(t)\mathbf{w}(t) 
\end{equation}
The function $x(t)$ represents the externally introduced change in the data point and is usually maintained as the identity. The data point undergoes dispersion that is scaled by $g(t)$ and described by the noise term $w(t)$. This forward process is commonly represented as a Markov Chain, with noise added at each time step based on a variance schedule ($\beta_1, ..., \beta_T$).

\begin{equation}
    q(x_{1:T} | x_0) = \prod\limits_{i=0}^T q(x_t | x_{t-1})
    \text{ ~with~ }
    q(x_{t} | x_{t-1}) = \mathcal{N}(x_{t-1}; \sqrt{1 - \beta_t} x_{t-1}, \beta_t \mathbf{I}). 
\end{equation}

Given the noisy state, we want the model to return the most probable, clean input image. So, for the backward process, we train a model to optimize the (variational lower bound of) the log-likelihood: 

\begin{equation}
    \mathbb{E}[ - \log p_\theta(x_0)] \leq \mathbb{E}\left[ -\log \left(p(x_T)) - \sum\limits_{t=1}^T \log\frac{p_\theta(x_{t-1} | x_t)}{q(x_t | x_{t-1})}\right)\right]
\end{equation}

The detailed derivations can be found in \cite{ddpm} and \cite{thermodynamics}.

For the reverse process, the conditional probability $p_\theta(x_{t-1} | x_t) := \mathcal{N}(x_{t-1}; \mu_\theta(x_t, t), \Sigma_\theta(x_t, t))$ is modelled as normal distribution and a neural network is optimized to predict $\mu_\theta$ and $\Sigma_\theta$.

\begin{figure}
    \centering
    \includegraphics[scale=0.7]{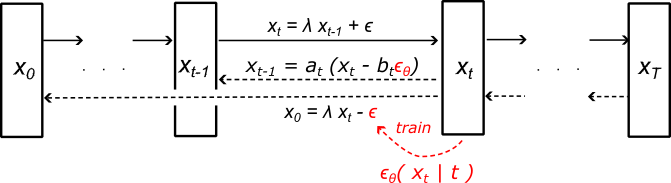}
    \caption{Reparametrization in sampling. The model does not predict the previous data point but the noise in relation to the clean image. The predicted noise and the diffusion process are used to interpolate between the clean image $x_0$ and the input $x_t$ to sample $x_{t-1}$ with the desired step size. $a_t$ and $b_t$ are functions of $t$ that encoder the stepsize and manage the interpolation between the clean and the noisy image.}
    \label{fig:intuition: reparametrized}
\end{figure}

Despite $\Sigma_\theta$ and the variance schedule $\beta_i$ can be learned, \cite{ddpm} opt for fixing all $\beta_t$ to a linear schedule to reduce computational costs. Specifically, they set $\Sigma_\theta(x_t | t) = \beta_t\mathbf{I}$ which allows to optimize solely for $\mu_\theta$. They observe that reparametrizing $\mu_\theta(x_t, t) = \frac{1}{\sqrt{1 - \beta}} \left(x_t - \frac{\beta_t}{\sqrt{1 - \prod_{i=1}^t (1 - \beta_t)}} \epsilon_\theta(x_t, t)\right)$ and optimizing for $\epsilon_\theta(x_t, t)$ yields even better performance. In addition, they suggest simplifying the loss by discarding some terms, again justifying this choice with better empirical performance. So, they end up optimizing the following objective; training to predict the noise in relation to the clean image (also see Figure~\ref{fig:intuition: reparametrized}):

\begin{equation}\label{eq:ddpm:simple_loss}
    \mathbb{E}_{x_0, \epsilon} \left[|| \epsilon - \epsilon_\theta(\sqrt{\prod_{i=1}^t (1 - \beta_t)} x_0 + \sqrt{1 - \prod_{i=1}^t (1 - \beta_t)} \epsilon, t)||^2 \right]
\end{equation}

During training, each gradient step involves independently sampling clean data points $x_0 \sim q(x_0)$, a random timesteps $t \sim \text{Uniform}({1, ..., T})$, and noise ($\epsilon \sim \mathcal{N}(\mathbf{0}, \mathbf{I})$).

Later research suggests potential improvements to the linear schedule \citep{ddpm2021improved}, and our experiments also demonstrate suboptimal performance, which highlights the inefficiency of sampling with this schedule.

\subsubsection{Stochastic Differential Equations}

Drawing from the same conceptual framework as in Langevin Dynamics, we formalize the diffusion process as a random phenomenon unfolding over time, which can be mathematically formulated as a Stochastic Differential Equation (SDE):

\begin{equation}
    d\textbf{x} = \textbf{x}(t) dt + g(t)d\mathbf{w}(t)
\end{equation}

This equation matches with the structure of the Langevin Dynamics Equation~\ref{eq:langevinddpm}, underscoring their similarity.

However, different from the discretized perspective of the Markov Chain, the reversion of a stochastic differential equation is represented by another stochastic differential process, expressed as:

\begin{equation}
    d\mathbf{x} = \left[\mathbf{x}(t) - g(t)^2 \nabla_{x(t)} \log p_t(x)\right] dt + g(t) d \mathbf{w}(t)
\end{equation}

When the score $\nabla_{x(t)} \log p_t(x)$ for all marginal distributions across time is known, we can effectively sample from this SDE. This score can be estimated through model training using score matching: A time-dependent model, denoted as $s_\theta(x, t)$, is trained to estimate $s_\theta(x(t), t)$, minimizing the following objective:

\begin{equation}
    \mathbb{E}_t \left\{\lambda(t) \mathbb{E}_{x(0)} \mathbb{E}_{x(t) | x(0)} \left[||s_\theta(x(t), t) - \nabla_{x(t)} \log q_{0t}(x(t)|x(0))||^2\right] \right\}.
\end{equation}

Estimating the score for the underlying ground truth distribution poses challenges, particularly in low-density regions with limited training samples. While adding noise to estimate scores is a valid approach, determining an optimal noise level for recovering the true distribution across the entire space is complex. Learning score functions over time mitigates this challenge.

At $t = T$, the data is standard normally distributed, simplifying score estimation. As time regresses $t \to 0$ and the data approaches the true underlying distribution, the accurate approximation of scores might be limited to high-density regions. However, in an iterative denoising process, all points would have already converged towards these high-density regions.

Note that ideally, we want to train the model to approximate $\nabla_{x(t)} \log q_{0t}(x(t))$, however with enough training data, one can show that this is equivalent to $\nabla_{x(t)} \log q_{0t}(x(t)|x(0))$. Additionally, note that for $x_t \sim \mathcal{N}(\mu_t x_0, \sigma_t^2)$, $x_t$ can be written as $\mu_t x_t + \sigma_t \epsilon$ and it holds that $\nabla_{x(t)} \log q_{0t}(x(t)) = - \frac{\epsilon}{\sigma}$. So, while in the approach motivated by Langevin Dynamics, we train to fit the noise $\epsilon$, in this approach motivated by SDEs, we optimize for the negative scaled noise $- \frac{\epsilon}{\sigma}$, which is yet again another reparametrization of the target.

Technically, this approach's main difference is its continuous nature and the possibility of optimizing it by solving the SDE. However, in practice, this approach is usually discretized for training (and sampling), and a neural network is used to approximate the score in the same way a network is trained to approximate the noise in the above approach.

\subsubsection{Stochastic Localization}

\citet{stochstic_loc} has recently drawn parallels between stochastic localization and the perspective of stochastic differential equations in diffusion models. Stochastic localization is a stochastic process where at each time step $t \in [0,\infty)$, we are given a random probability measure $\mu_t$.
As time progresses $(t \to \infty)$, the probability measure $\mu_t$ localizes, that is, it converges to a point $\mu_t \to \delta_{x_*}$, where $x_*$ is a random variable. The only requirement is that this process must be martingale. 
This means that at a particular time, the conditional expectation of the next value in the sequence is equal to the present value, regardless of all prior values. As with the previous methods, the general idea is that if we can construct this process, we can sample from $\delta^*$. 

Let $\mathbf{Y}_t$ be such a process, and for simplicity, assume it follows a Gaussian distribution:

\begin{equation}
    \mathbf{Y}_t = tx_* + \mathbf{W}_t
\end{equation}

where $W_{t \geq 0}$ is a Wiener process. We observe that, as time $t$ increases, the signal-to-noise ratio also increases. \citet{stochstic_loc} show that this process is the unique solution to a stochastic differential equation, coinciding with the one derived in \citet{song2021scorebased}. This gives rise to another mathematical framework to analyze the properties of diffusion processes and models.

\section{Diffusion Models in Discrete State Space}

The diffusion process has been successfully adapted to various spaces, such as discrete state spaces \cite{austin2021structured} and function spaces \cite{lim2023function}. In graphs, the former adaptation can be deployed \cite{graphsdiscrete2022diffusion}, \cite{digress}.

While certain adjustments are necessary, the underlying concept remains the same. The approach involves diffusing clean input graphs until they resemble random graphs and then learning to reverse this process. The diffusion and sampling processes must work in the discrete state space. Each datapoint $x$ is expressed as a one-hot encoding, assuming one of $d$ states: $x \in \{0, 1\}^d$. The noise is characterized by transition matrices $Q^1, ... Q^t$, where $[Q^t]_{ij}$ is the probability of transitioning from state $i$ to state $j$: $q(x^t | x^{t-1}) = x^{t-1} Q^t$.

Based on this representation, one can derive the marginal and posterior distribution for $t$ steps: 

\begin{equation}\label{eq:discrete:marginal}
    q(x_t ~|~ x_0) = x_0 \mathbf{\bar{Q}}^t \text{ with } \bar{\mathbf{Q}}^t = \mathbf{Q}^1\mathbf{Q}^2 ... \mathbf{Q}^t 
\end{equation}

and

\begin{equation}\label{eq:discrete:posterior}
    q(x_{t-1} ~|~ x_t, x_0) = \frac{x_t (\mathbf{Q}^\texttt{T}) \bigodot x_0 \mathbf{\bar{Q}}^{t-1}}{x_0 \mathbf{\bar{Q}}^tx_t^\texttt{T}}
\end{equation}

Now, one can train to directly predict the logits $p_\theta(x_{t-1} ~|~ x_t)$. However, many approaches opt for a sampling procedure, wherein the model predicts the clean input $p_\theta(x_0 ~|~ x_t)$, uses renoising $q(x_{t-1} ~|~ x_t, x_0)$, and marginalizing over the one-hot encodings:

\begin{equation}
    p_\theta(x_{t-1} ~|~ x_t) \propto \sum\limits_{x_0} q(x_{t-1}, x_t ~|~ x_0) p_\theta(x_0 ~|~ x_t)
\end{equation}

\subsection{Sampling and the approximated function}

The whole pipeline of denoising diffusion models has three parts that all work together to generate new samples. The diffusion process is the iterative process of adding small (random) perturbations to the data, which is used to generate training data. 

The denoising part of a denoising diffusion pipeline is the sampling process. This process is based on parts of the target distribution that are unknown, intractable, or unfeasible to compute. So, one part of the sampling is a function that is approximated by a (graph) neural network. The exact sampling procedure and the approximated function rely on each other. Depending on the sampling strategies, different objectives are optimized, and the chosen neural network approximates different functions. We want to understand the role of the three components and try to disentangle their influence.

Graph diffusion, as presented in \cite{digress}, is based on the algorithms presented in \cite{ddpm}.

Given a graph $G = \{X, E\}$, as described in the discrete setting above, the state of each node and edge of the graph is encoded as a one-hot vector. A node $x$ can take $d$ states $x \in \{0, 1\}^d, X\in \{0, 1\}^{n \times d}$. Analogously for each edge. 

The marginal and posterior distributions are given by Equation~\ref{eq:discrete:marginal} and Equation~\ref{eq:discrete:posterior}.

A graph neural network is trained to solve a classification task on each node and edge, given a noisy graph $G^t = \{X^t, E^t\}$. It optimizes the cross-entropy between the predicted probabilities $\hat{p} = (\hat{p}^X, \hat{p}^E)$ for each node and edge and the true graph:

\begin{equation}
    \sum\limits_{i=1}^n \text{cross-entropy}(x_i, \hat{p}_i^X) + \lambda \sum\limits_{i,j=1}^n \text{cross-entropy}(e_{ij}, \hat{p}_{ij}^E)
\end{equation}

Once trained, one samples from the reverse process 
$$p_\theta(G^{t-1} ~|~ G^t) = \prod\limits_{i=0}^n p_\theta(X_{i:}^{t-1} ~|~ G^t) \prod\limits_{i,j=0}^n p_\theta(E_{ij:}^{t-1} ~|~ G^t),$$ 
which can be estimated from the network predictions: 

\begin{equation}
    p_\theta(x_{t-1} ~|~ G^t) = \sum\limits_{x \in \mathcal{X}} p_\theta(x_{t-1} ~|~ x_0 = x, G^t) \hat{p}^X(x)
\end{equation}

where 

\begin{equation}
    p_\theta(x_{t-1} ~|~ x_0 = x, G^t) = \begin{cases}
        q(x_{t-1} ~|~ x = x_0, x_t) & \text{if } q(x_t ~|~ x = x_{t-1}) > 0\\
        0 & \text{otherwise}
    \end{cases}
\end{equation}

As is done for images, where the diffusion is applied to each pixel independently, the diffusion process is not defined on graphs but independently on edges and nodes. The structural information of the graph is neglected in this step. Instead of a standard neural network that approximates the gradient of points in the data, Digress uses a graph neural network to solve a classification task on each node and edge. As the sampling also uses the information from the diffusion process, the graph structure is only considered in the learned weights of the graph neural network.

This raises questions about to what extent the diffusion, the graph neural network, or the sampling contribute to good-quality samples. In their work, they suggest using the marginal distribution of classes in the training data and show superior sampling quality when using this process. This indicates that the noise process and the information put into it significantly affect the sampling quality. 

Other works on images, as \cite{bansal2024cold}, claim noise is unnecessary, showing high-quality samples for deterministic diffusion processes.

Several questions arise considering the influence of certain parts of the algorithms pipeline and their respective biases.

{\bf Q1}: What is the role of noise in the diffusion denoising pipeline, and do we need it at all? We investigate the importance of the different parts in simulations and give some insights into their role. 

{\bf Q2}: How much influence does the sampling have on the performance? Some works suggest reparametrization in the sampling. While approaches on graphs train to predict the clean graph, other works such as \cite{ddpm} note that predicting the clean image is less accurate.

{\bf Q3}: What does the neural network approximate, and do we need the complexity? When solving the "simple" classification task for the graph setting, could a simpler model achieve similar results? How much structural information do we introduce by the iterative sampling procedure, including the forward noise process?

How to approach these questions is not ad hoc clear, and the complexity of graphs and graph neural networks introduce an additional degree of complexity. As a starting point, we want to investigate the three components in a much simpler setting. This helps to break it down into a setting that we can visualize and allows us to build intuition in a more graspable setting.

\section{Diffusion and denoising in a simple setting.}

\subsubsection{Setup}

Consider a set of points in two dimensions originating from some unknown distribution $p$. We want to generate new samples $ x \sim p$ from this distribution. We cannot sample from it because we cannot access the underlying distribution. However, we can train a denoising diffusion model to sample from an approximated distribution $\tilde{p}$.

For a simple analysis, we choose a mixture of two Gaussians. Figure~\ref{fig:sim:trainingdata} shows the density and the score of the chosen ground truth distribution with $$\mu_1 = (-4, -4), \mu_2 = (4, 4), \sigma_1 = \begin{pmatrix}0.3 & 0 \\ 0 & 0.1\end{pmatrix} \text{ and } \sigma_2 = \begin{pmatrix}0.2 & 0 \\ 0 & 0.2\end{pmatrix}.$$ 

Our simulations (ref Section~\ref{sec:simulations} suggest the following answers to the questions raised in the section above:

{\bf (A1) We do not {\sl need} the noise.} 
\cite{song2019genmodelscore} observe that the gradient approximation is poor in low-density regions of the data and address the problem by adding noise. If we do not introduce any perturbation, we only sufficiently approximate the data gradient close to high-density regions. Noise mitigates the problem by diffusing the training points and leaving no low-density regions. Clearly, too much noise leaves no signal. Hence, the amount of noise added is crucial. However, by iteratively adding tiny perturbations and learning an iterative backward process, we can approximate the time-dependent ground truth distribution even when starting far away from high-density regions. However, the conclusion that we need noise in the sense of randomness is misleading. As long as we manage to cover the space sufficiently, the diffusion process can also be of a deterministic nature. We show experiments on that in Section~\ref{sec:simulations}.

{\bf (A2) Diffusion schedule and sampling process are crucial for the performance.}

Unsurprisingly, the diffusion schedule plays an essential role in the proper approximation of the reverse process. Figure~\ref{fig:schedule} visualizes the influence of different schedules for $\beta, \alpha$ and $\Bar{\alpha}$. The linear schedule leads to faster convergence to a standard normal distribution and thus loses much signal in the first steps. As a result, the later timesteps contain little to no signal and are worthless for training. The cosine schedule results in a smoother transition; thus, later timesteps contain a more valuable signal for the training process. As all timesteps are equally likely to be sampled during training, lower, smoother diffusion is better.

In addition, what exactly is approximated by the neural network significantly influences the performance. Both the distribution and its likelihood follow mathematical rules that are hard to enforce with a neural network. Thus, predicting the likelihood of a data point is challenging. While it is only a reparametrization of the target, locally approximating the score of the likelihood allows the inclusion of additional information about the noise process, seems more accessible, and empirically results in better performance.

{\bf (A3) A simple network approximates the data distribution reasonably well.}

The network only partially approximates the distribution's score. However, even though our network's architecture is simple, and thus, its approximation power is limited, we learn essential features of the ground truth distribution in all three settings.

It is impossible to learn random independent noise. So, the model does not approximate the actual reverse process but the gradient of the distribution in each step. For each point, the model learns a mapping that moves every point closer to a high-density region of the training data.

Aligned with the intuition behind stochastic localization, the network learns the gradient of the distribution for each time step.

\subsection{Simulations}\label{sec:simulations}

\subsubsection{Generation process and experimental setup}

\begin{figure}[t]
    \centering
    \includegraphics[width=0.4\textwidth]{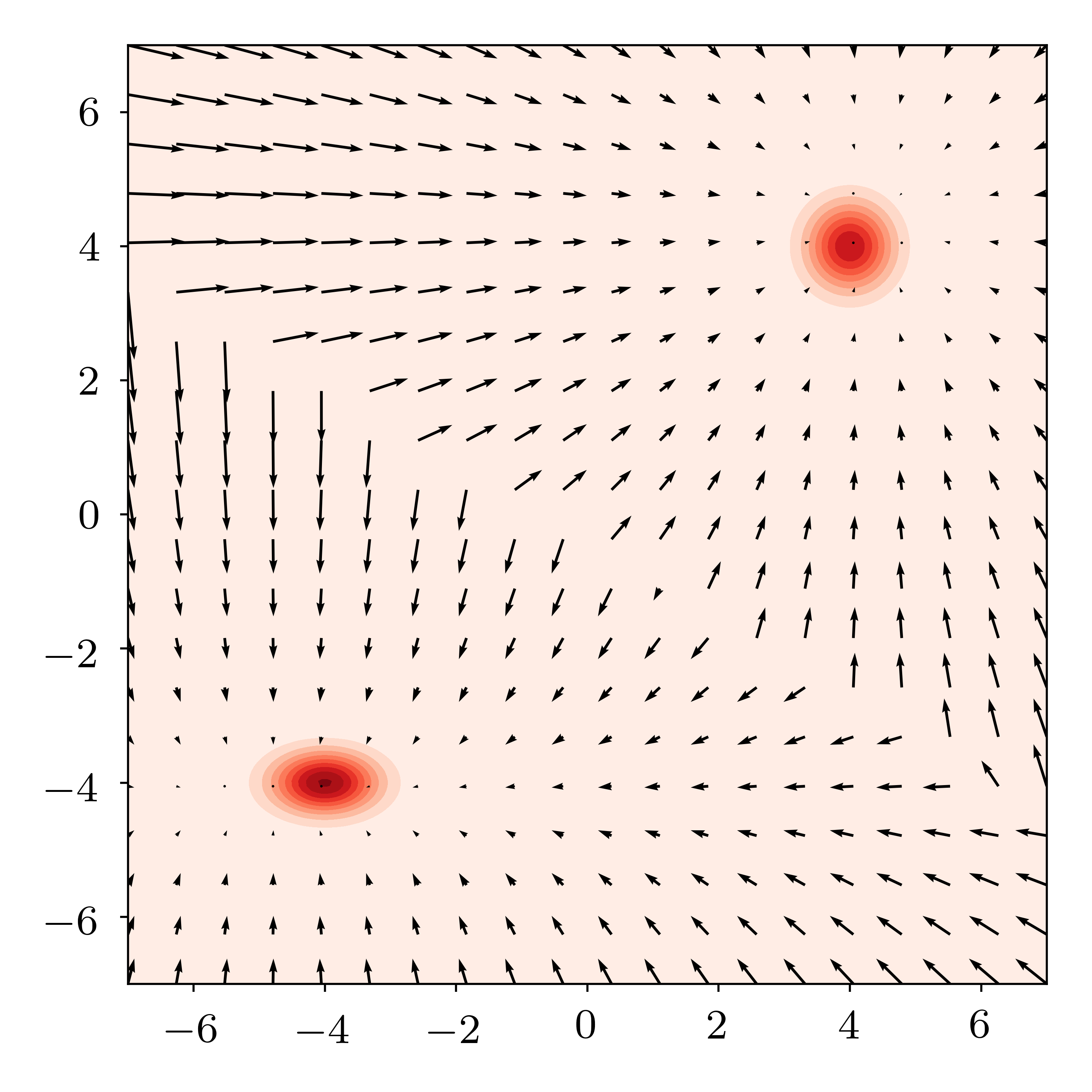}\hfill
    \includegraphics[width=0.4\textwidth]{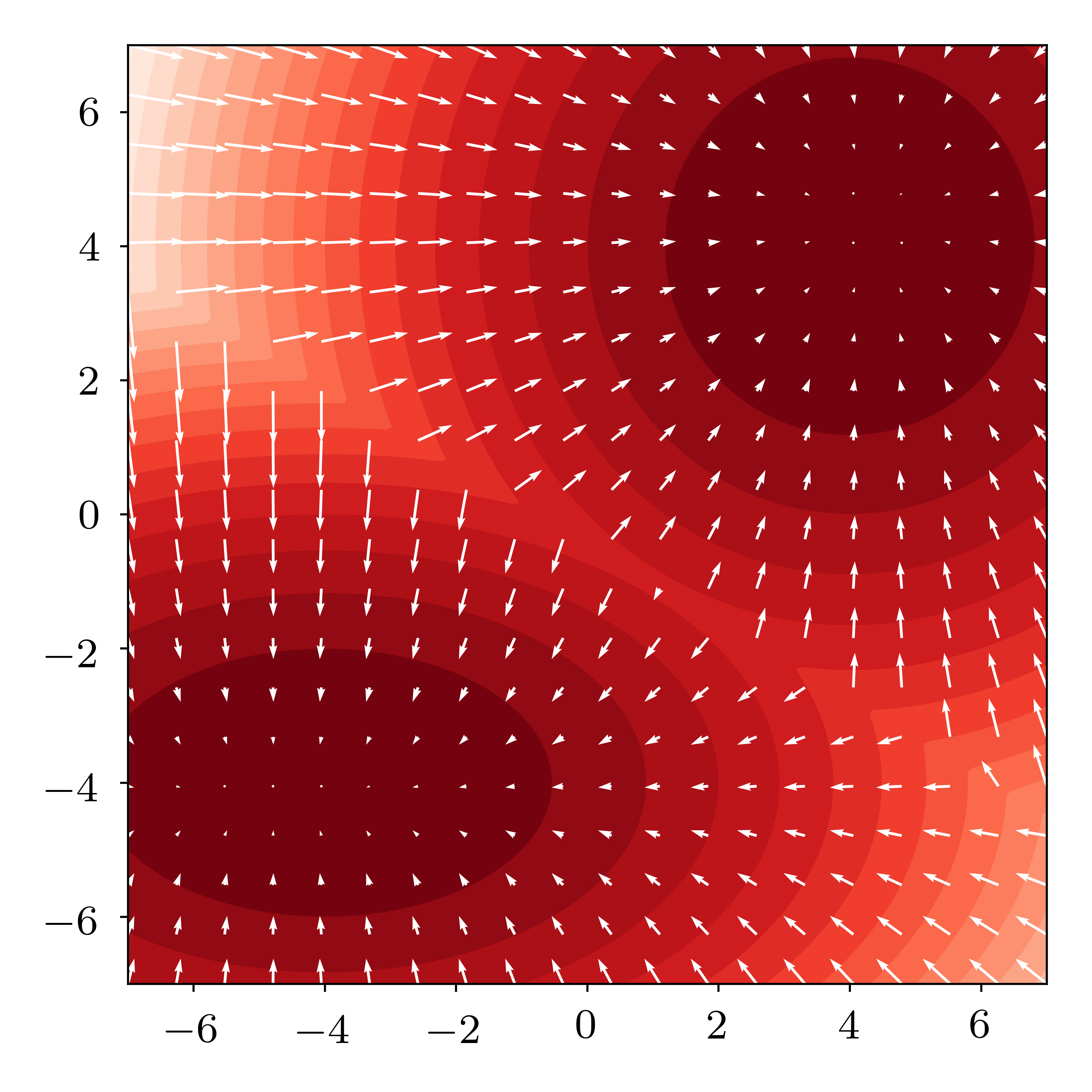}
    \caption{Ground truth data distribution used to sample training points. The left figure shows the density, and the right figure shows the log-likelihood. The arrows indicate the direction of the score $\nabla_x \log p(x)$.}
    \label{fig:sim:trainingdata}
\end{figure}

We generate data from a mixture of two Gaussians. We randomly sample 5,000 points from each of the two distributions, so 10,000 training points overall. The distribution we sample the training data from is visualized in Figure~\ref{fig:sim:trainingdata}. 

\begin{figure}[t]
    \centering
    \includegraphics[scale=0.7]{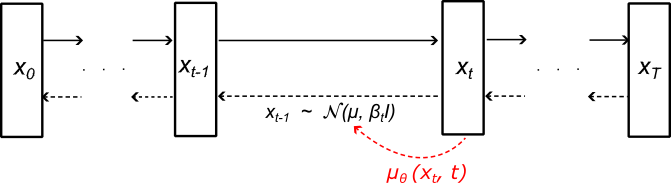}\\
    a)\\
    \includegraphics[scale=0.7]{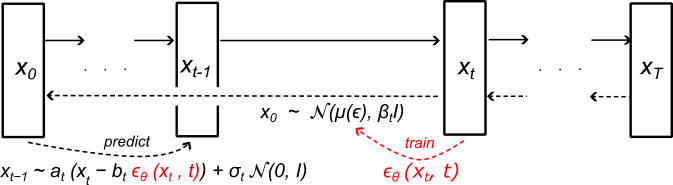}\\
    b)\\
    \includegraphics[scale=0.7]{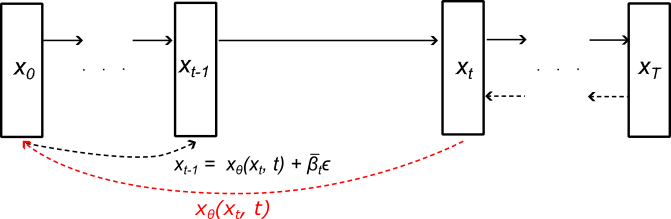}\\
    c)\\
    \caption{Visualization of the three investigated sampling methods. Red indicates the part that the model predicts.}
    \label{fig:sim:sampling}
\end{figure}

For each of the investigated sampling methods (see Figure~\ref{fig:sim:sampling}), we train the same neural network architecture: a simple multi-layer perceptron with two relu layers of width 20 and a final linear layer as output.

The model is trained with a batch size of 64 for 50 epochs using the Adam Optimizer from PyTorch. We note that we did not tune the neural network in any way and used the same architecture and hyperparameters for the three different tasks. Given the simplicity of the chosen problem, a comparison on this basis is still fair and justified.

\subsubsection{Different Noise Schedules}

\begin{figure}
    \centering
    \begin{subfigure}[b]{0.3\textwidth}
         \centering
         \includegraphics[width=\textwidth]{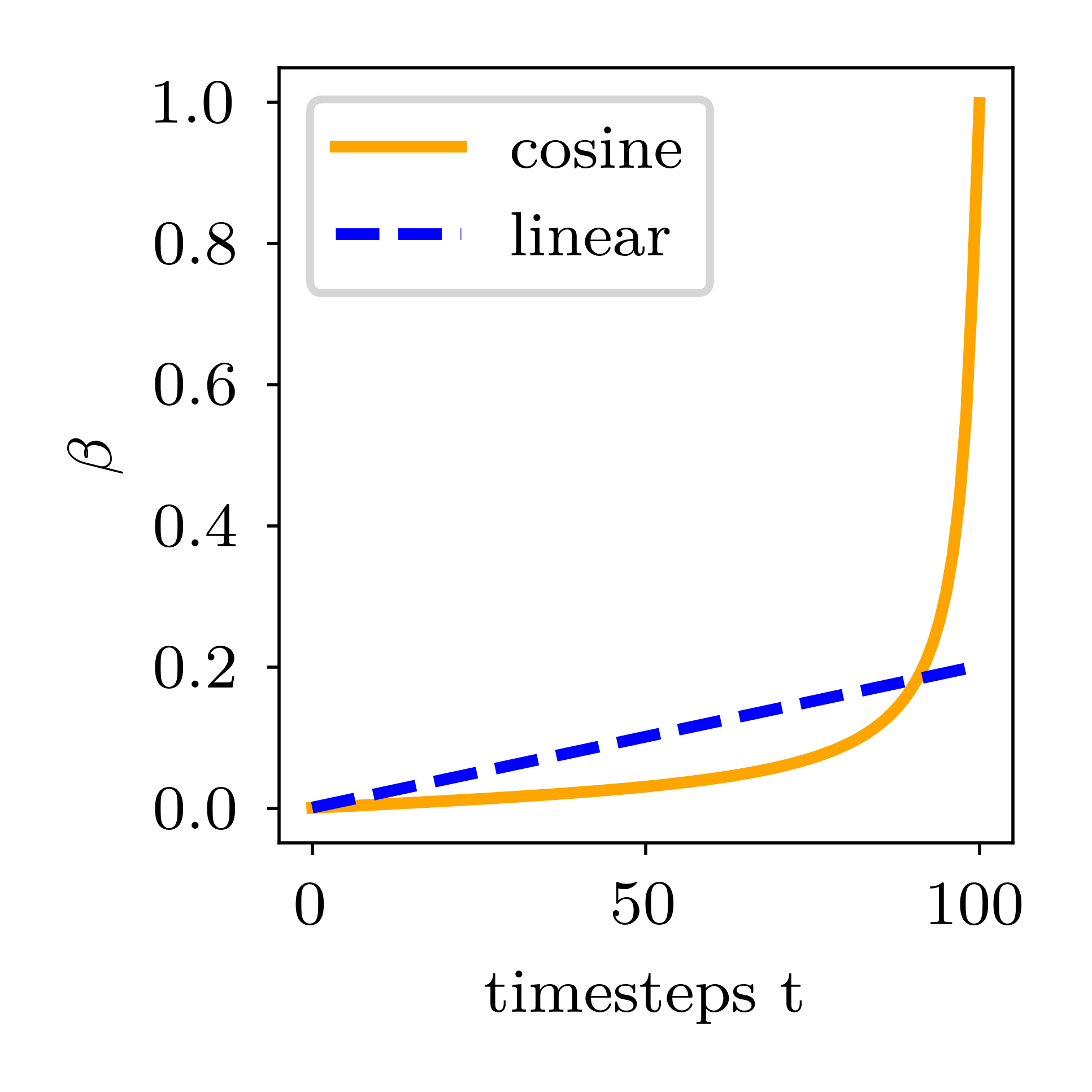}
         \caption{$\beta_t$}
         \label{fig:schedule:beta}
    \end{subfigure}
    \begin{subfigure}[b]{0.3\textwidth}
         \centering
         \includegraphics[width=\textwidth]{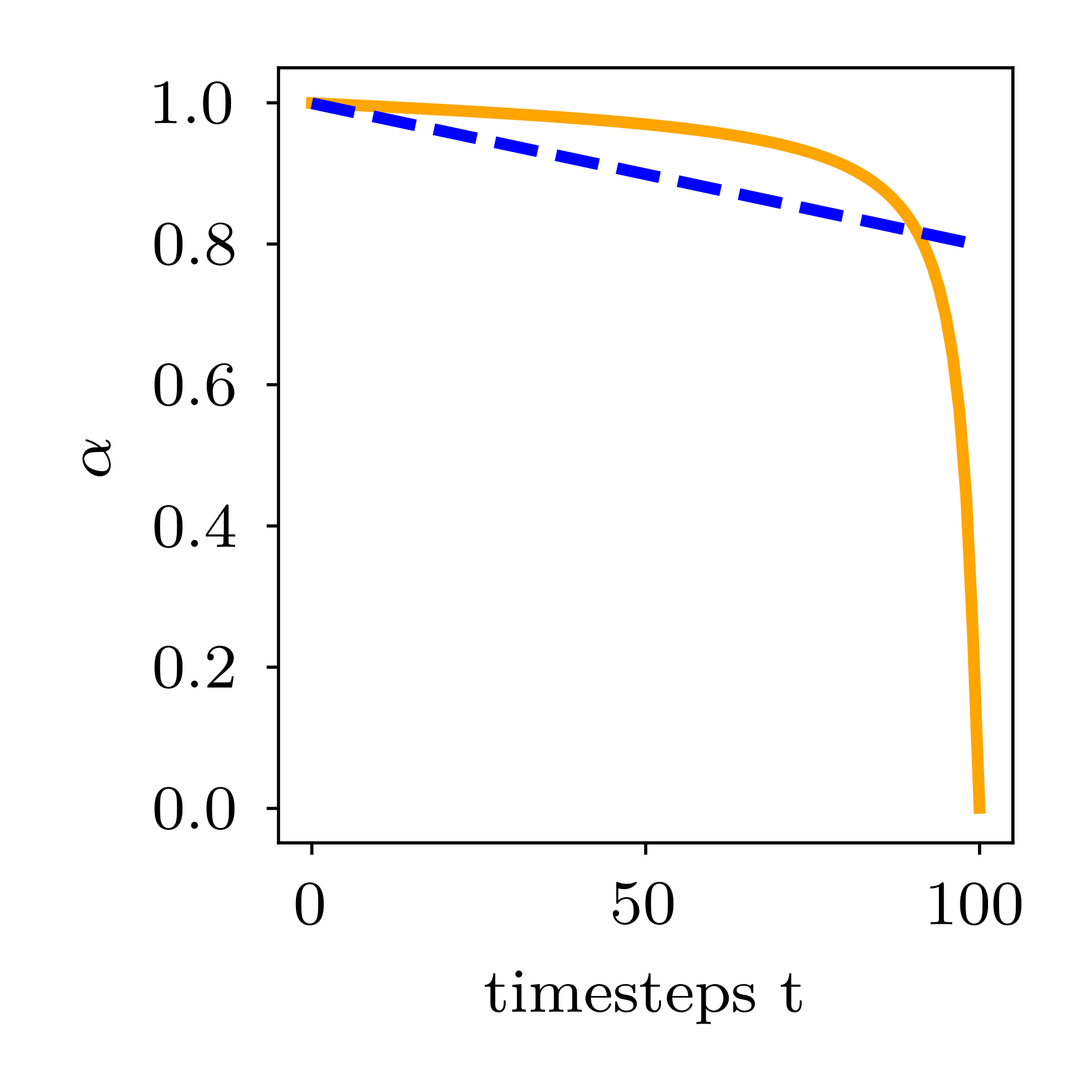}
         \caption{$\alpha_t = 1 - \beta_t$}
         \label{fig:schedule:alpha}
    \end{subfigure}
    \begin{subfigure}[b]{0.3\textwidth}
         \centering
         \includegraphics[width=\textwidth]{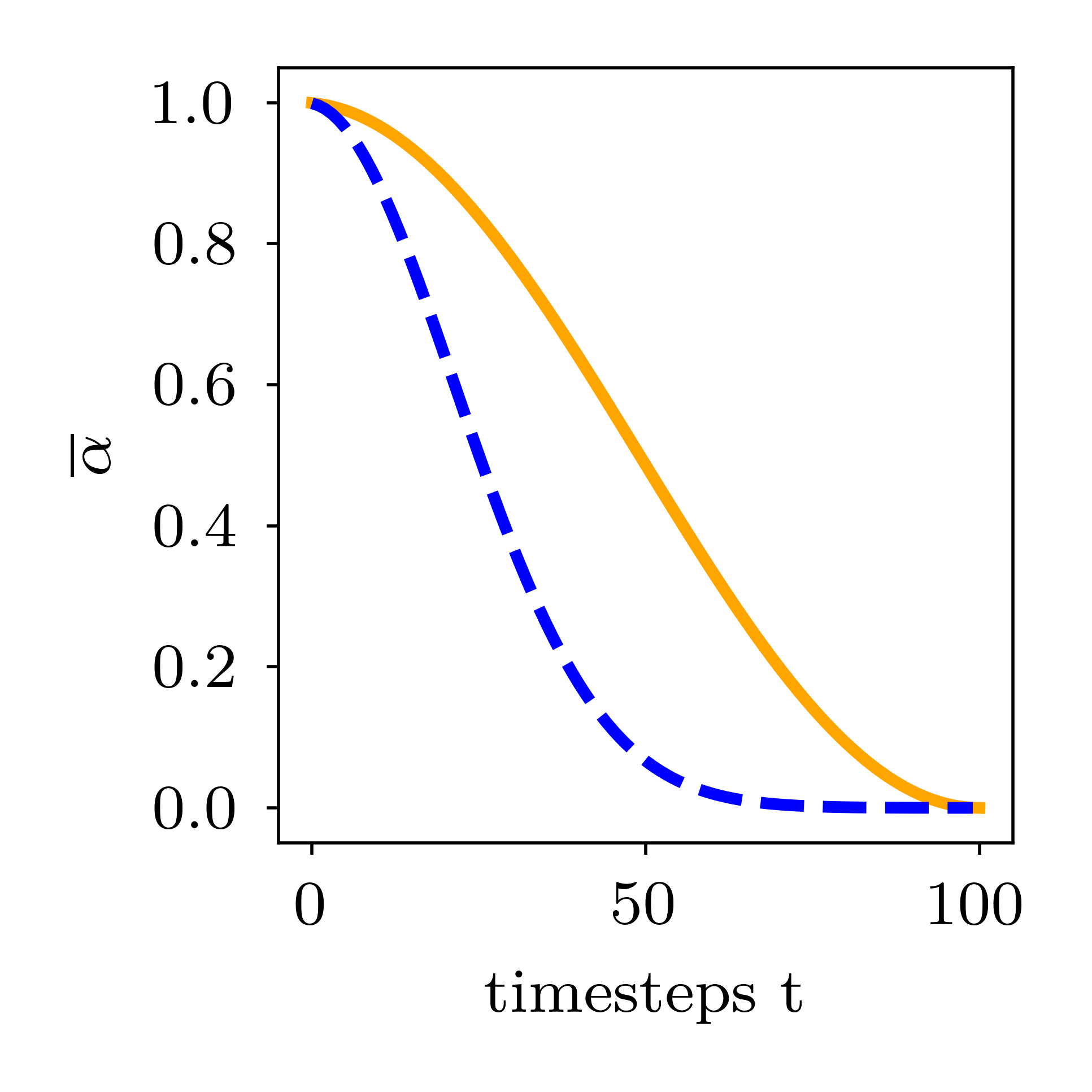}
         \caption{$\Bar{\alpha}_t = \prod_{i=1}^t \alpha_t$}
         \label{fig:schedule:alphabar}
    \end{subfigure}\par\bigskip
    \begin{subfigure}{\textwidth}
        \includegraphics[width=0.19\textwidth]{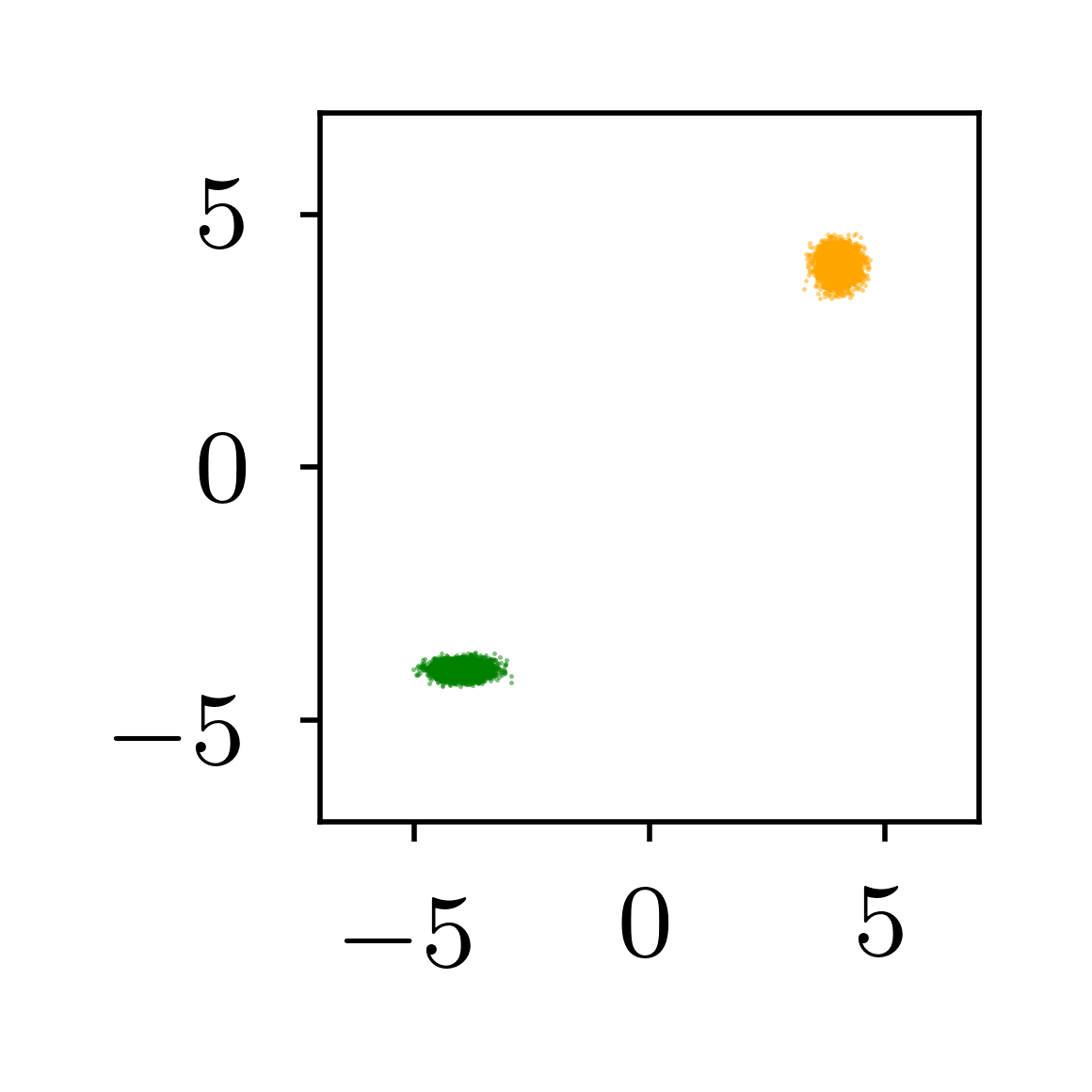}
        \includegraphics[width=0.19\textwidth]{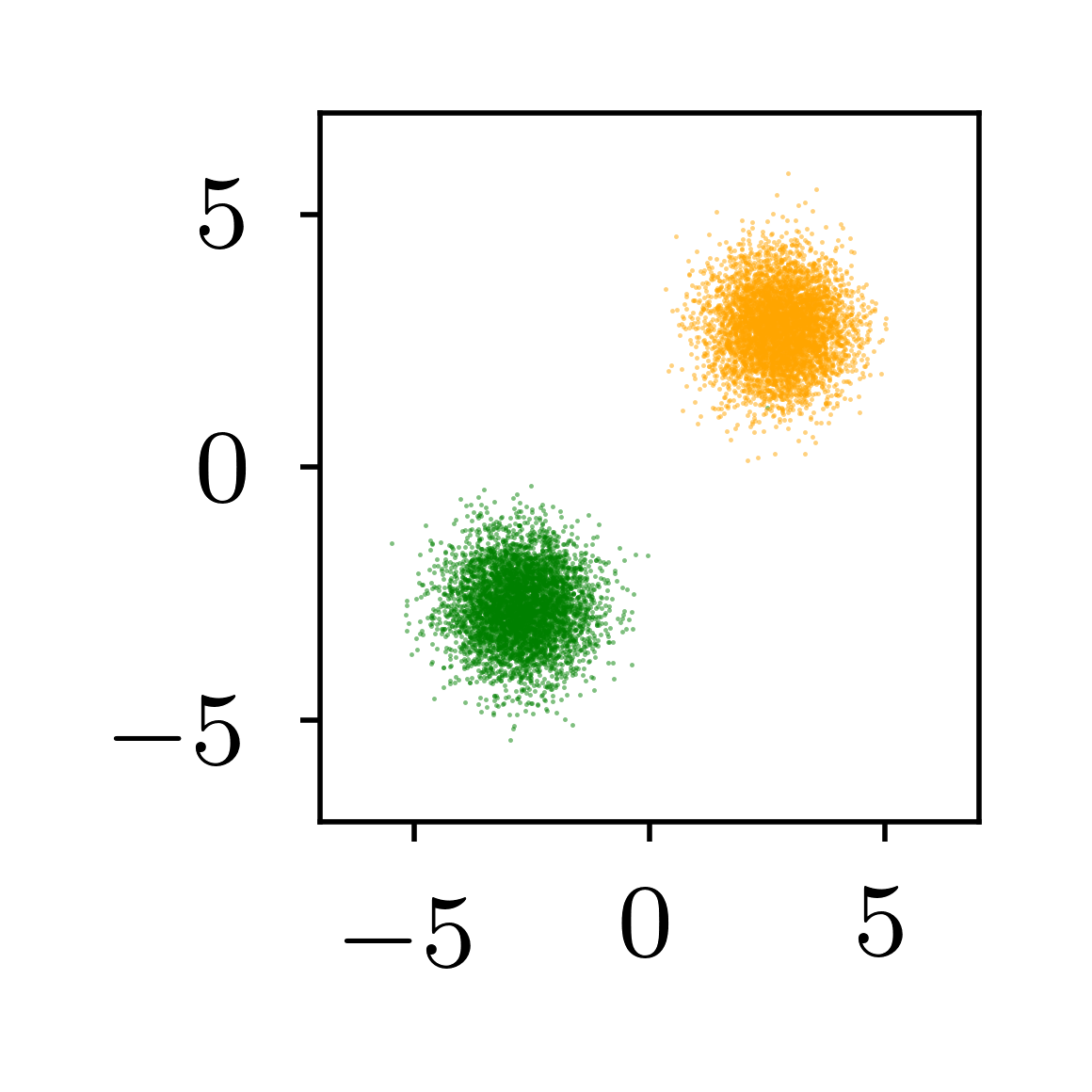}
        \includegraphics[width=0.19\textwidth]{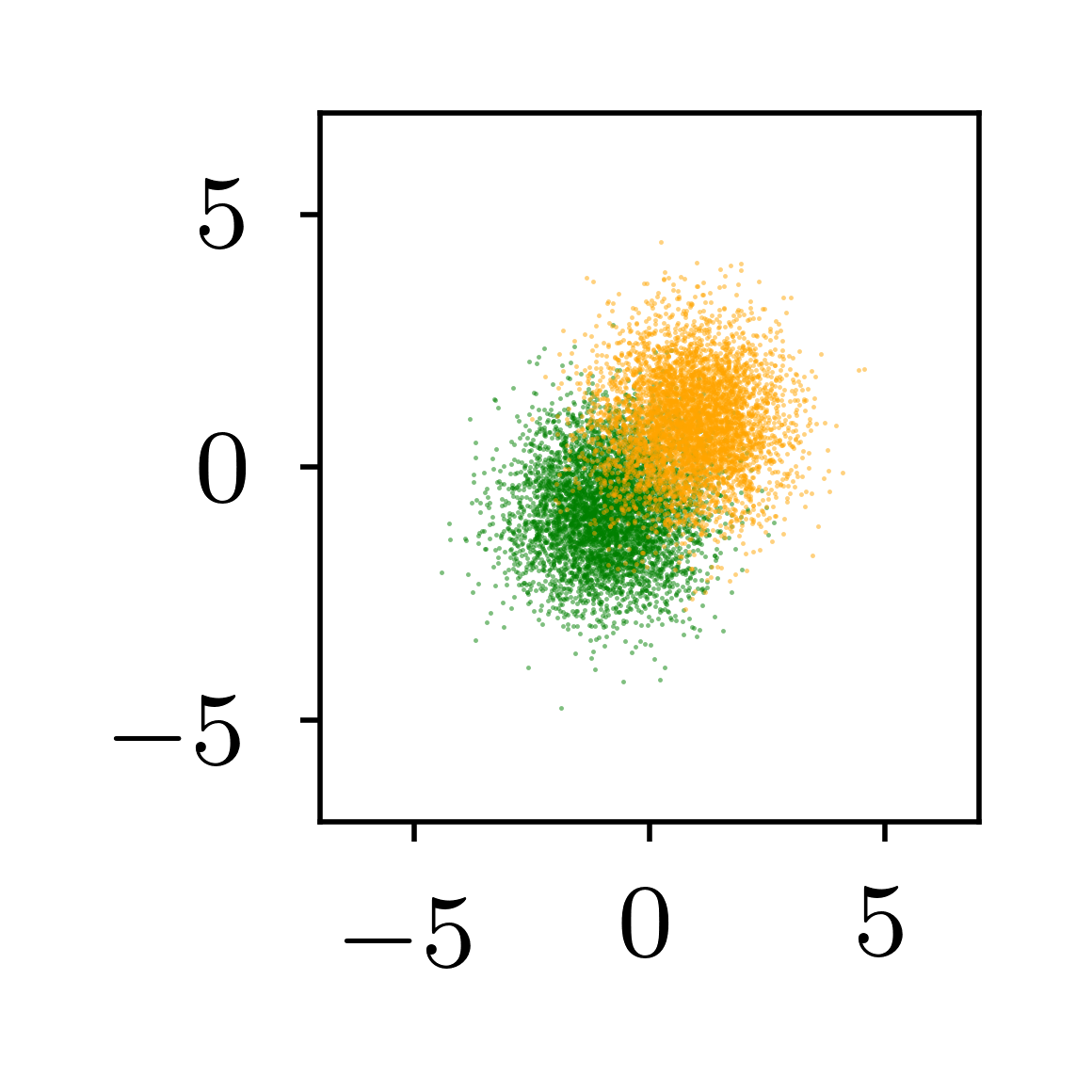}
        \includegraphics[width=0.19\textwidth]{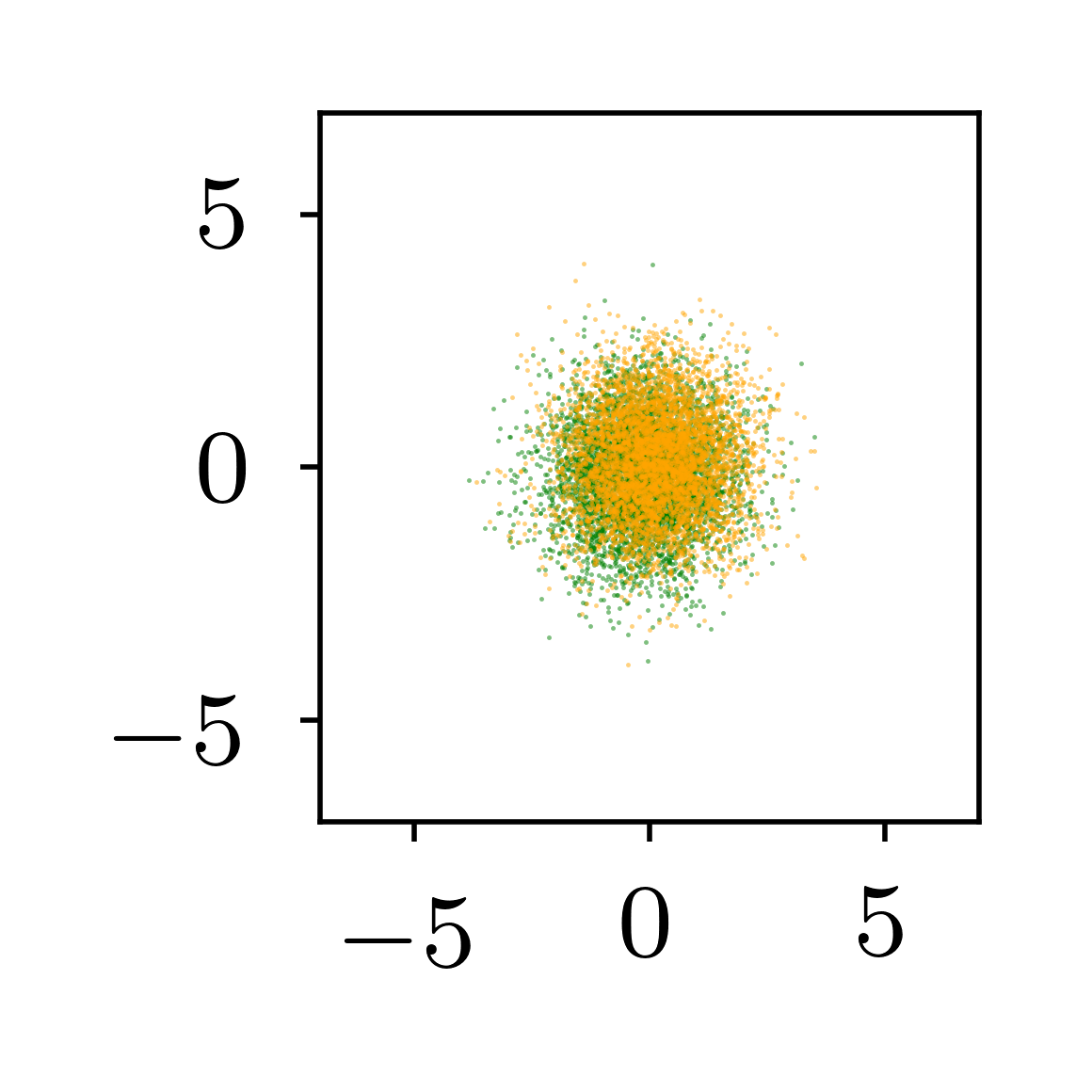}
        \includegraphics[width=0.19\textwidth]{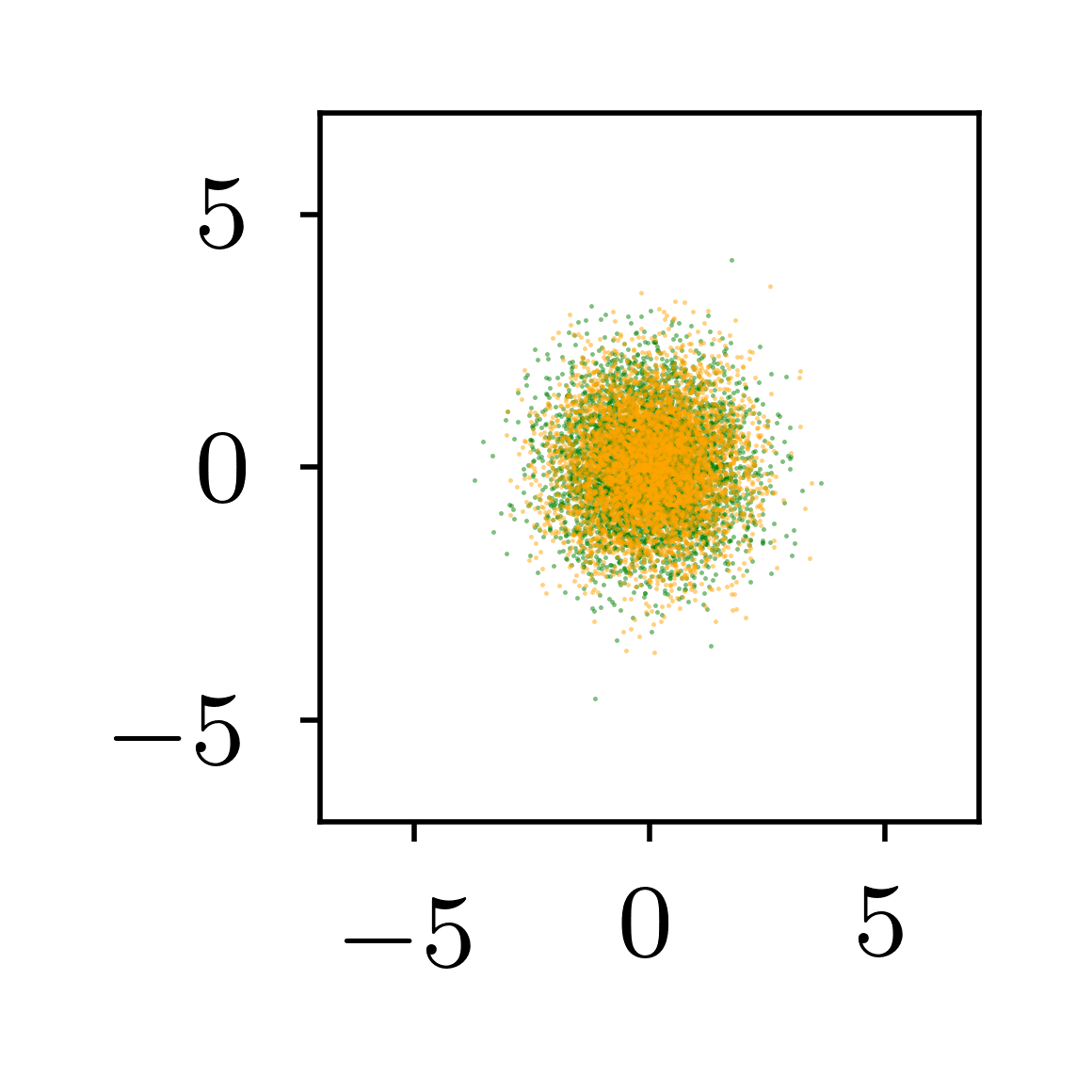}    
        \caption{a linear diffusion process with normal noise}
         \label{fig:schedule:linear}
    \end{subfigure}\\
    \begin{subfigure}{\textwidth}
        \includegraphics[width=0.19\textwidth]{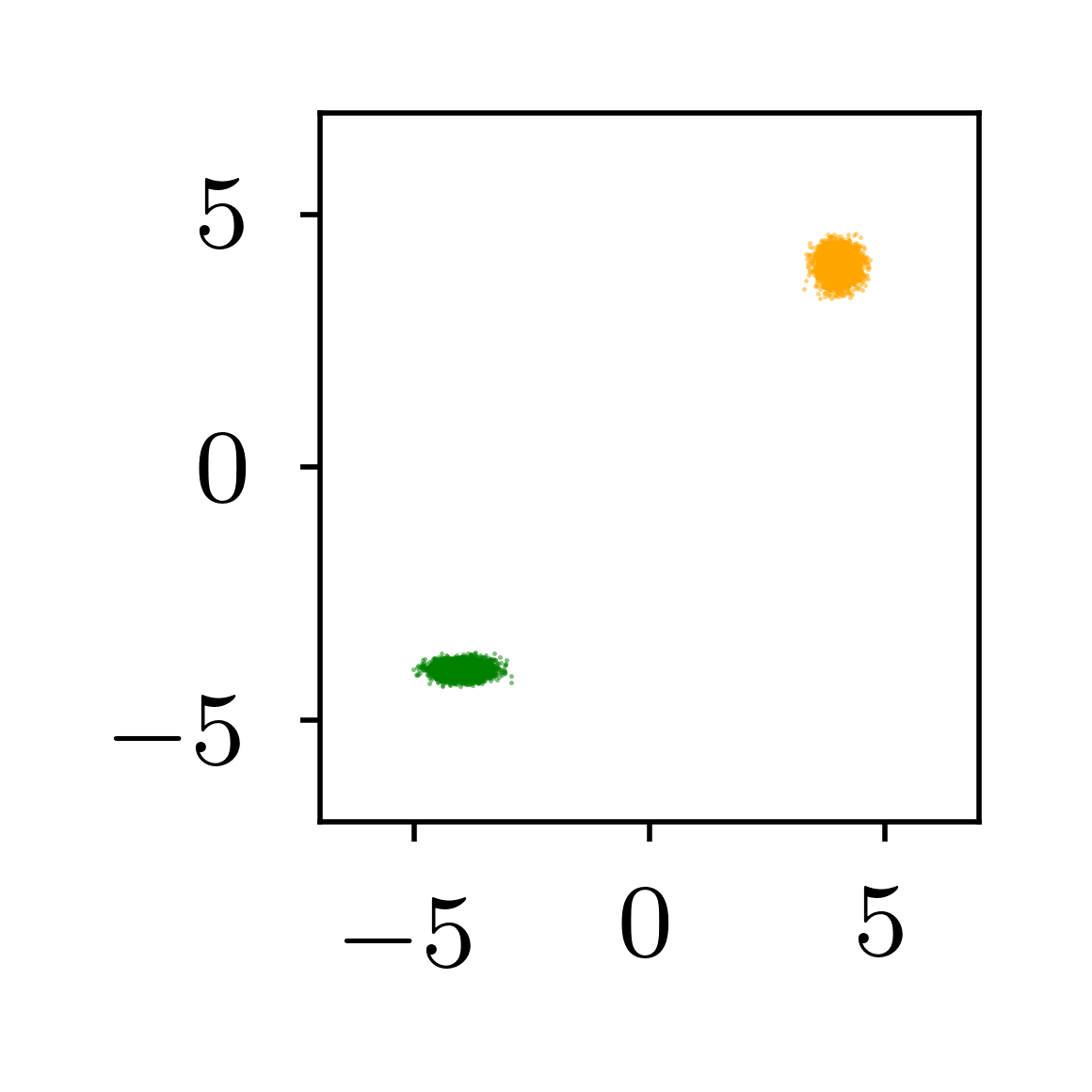}
        \includegraphics[width=0.19\textwidth]{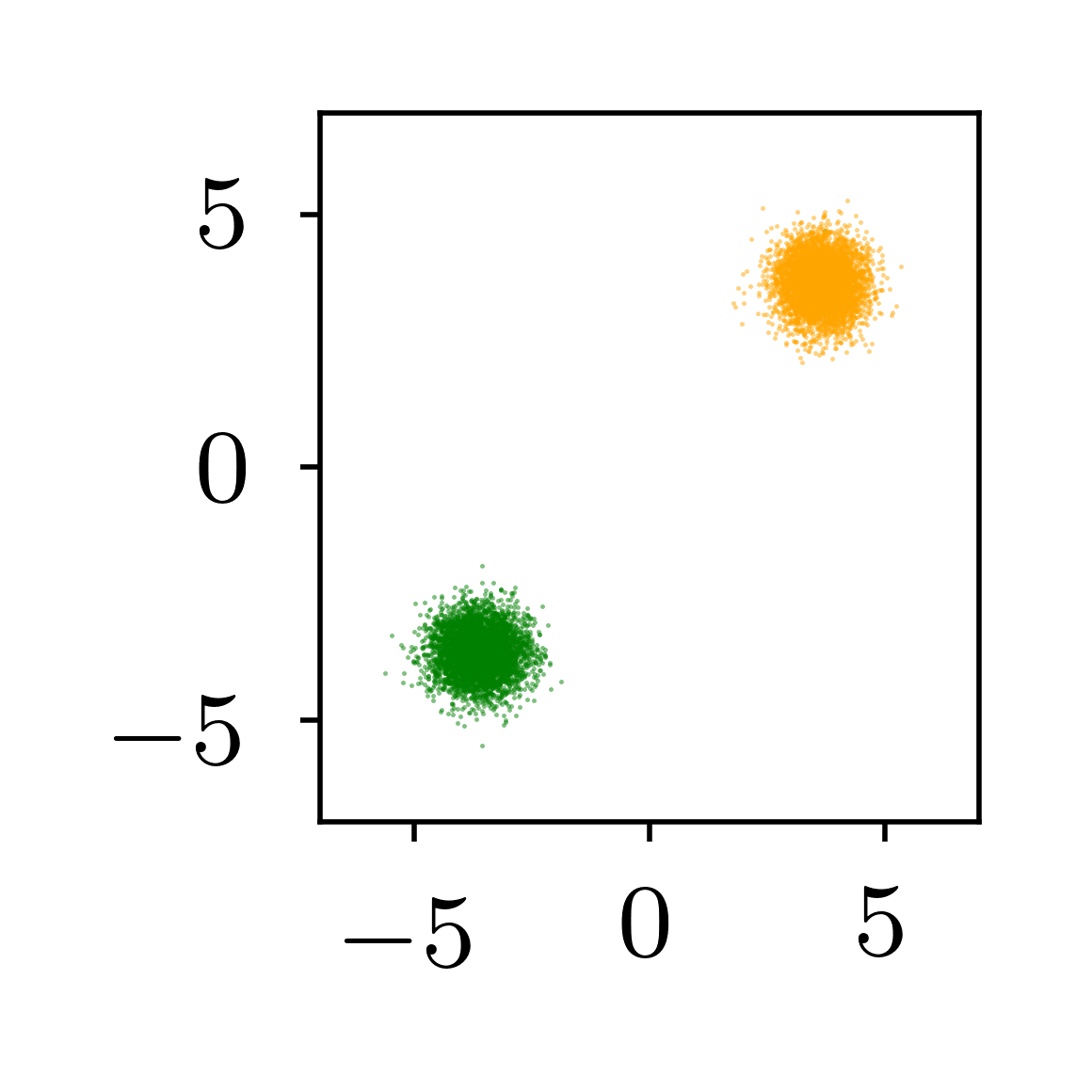}
        \includegraphics[width=0.19\textwidth]{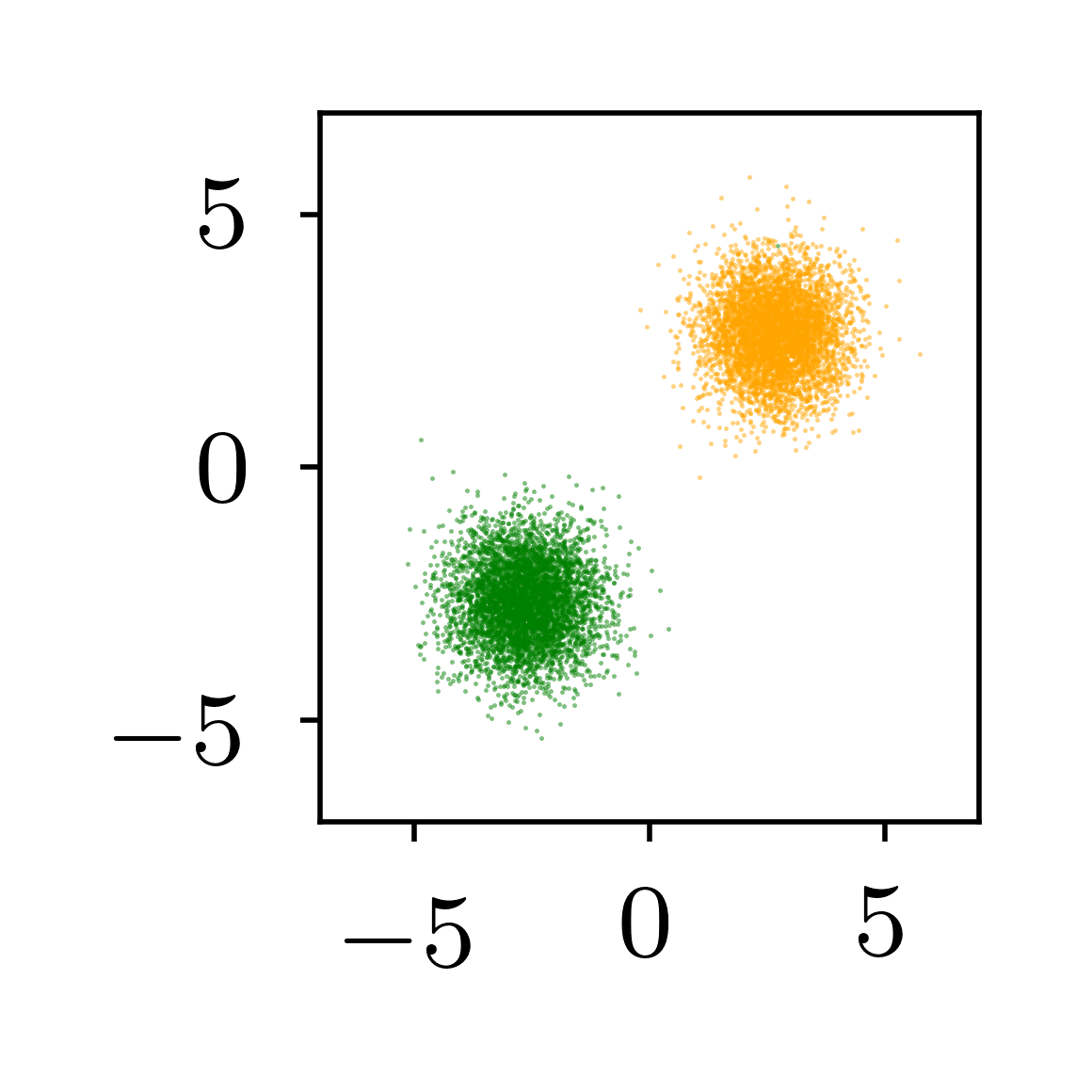}
        \includegraphics[width=0.19\textwidth]{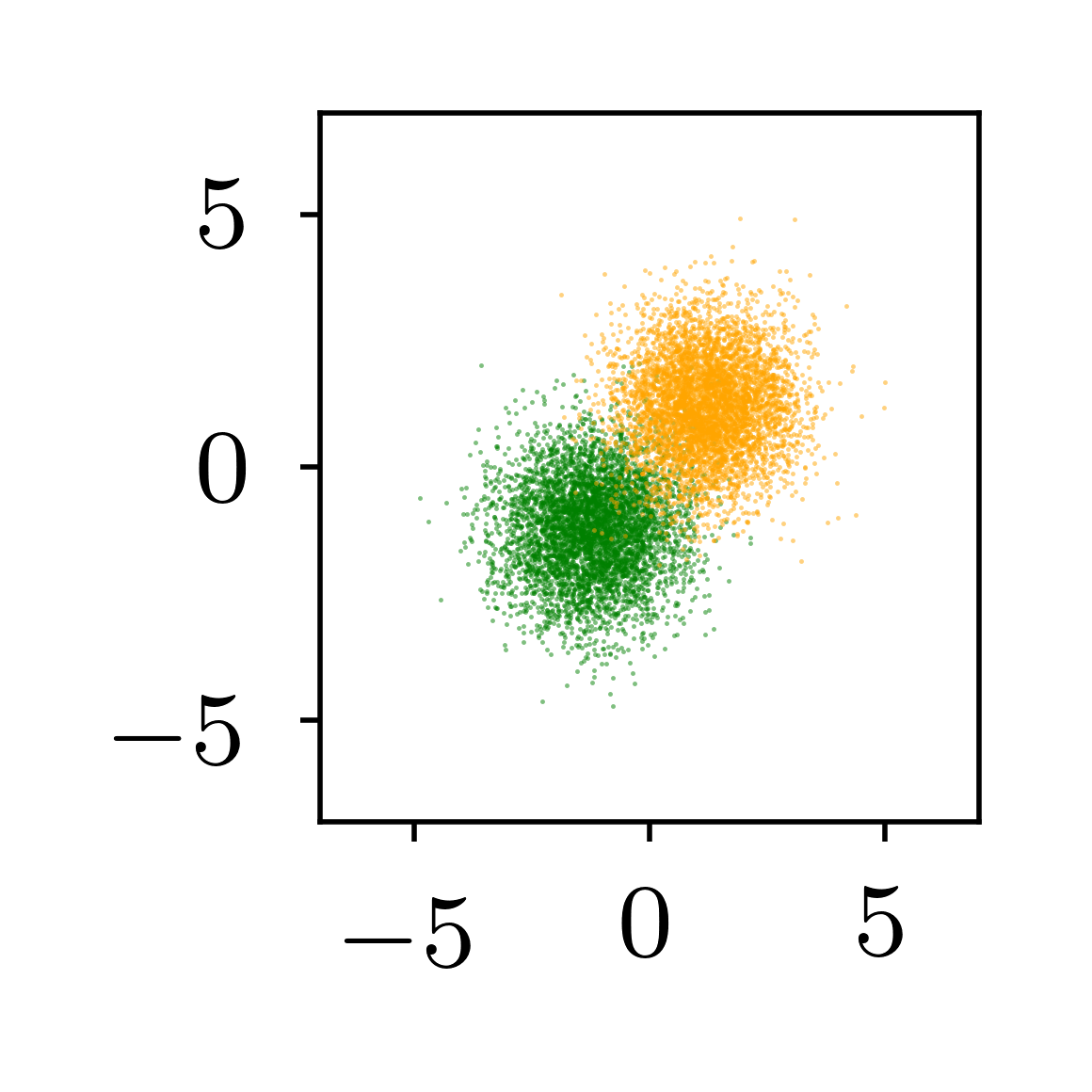}
        \includegraphics[width=0.19\textwidth]{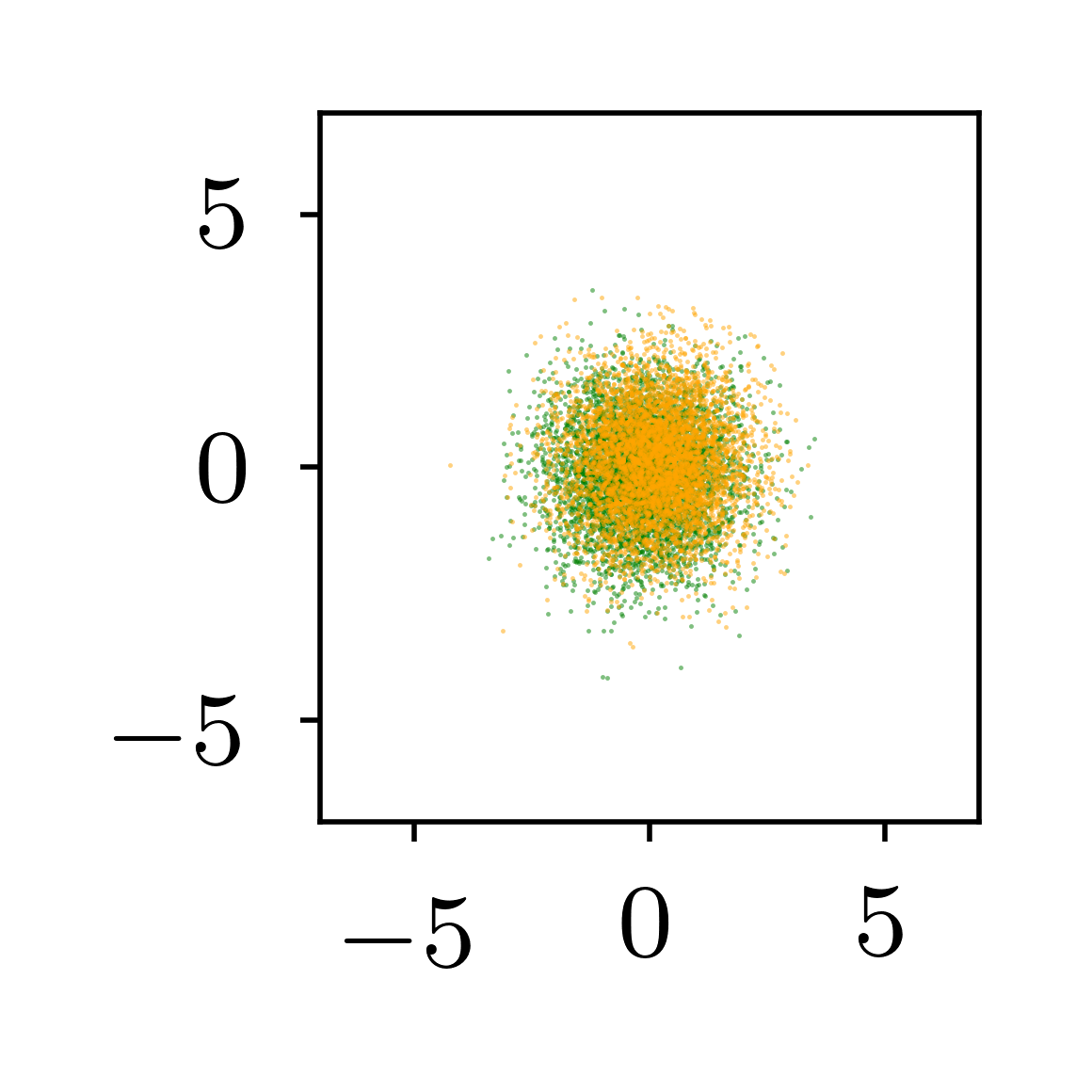}
        \caption{a cosine diffusion process with normal noise}
         \label{fig:schedule:cosine}
    \end{subfigure}\par\bigskip
    \caption{The noise schedule makes a difference. For linear diffusion, most information is lost in the early time steps, and later steps hold little to no information about either the original distribution or the diffusion process. Controlled by $\Bar{\alpha}$, the information in the cosine diffusion process degrades slower, so later steps still hold valuable transition information for the training. Visualizations of the diffusion process in Figures~(d) and~(e) show timesteps t = 0, 27, 54, 81, 99 from left to right.}
    \label{fig:schedule}
\end{figure}

Different papers observe that the noise schedule in the training can play a crucial role in the performance of the generative model. While the original work of ~\citet{ddpm} suggests a linear schedule, recent works usually use the cosine schedule introduced in~\citet{ddpm2021improved}. Empirically, the latter proves to yield better performance. In the linear case, a lot of the time, steps fall into the range where the data is indistinguishable from random noise. In those steps, the training data does not hold enough information for learning. The cosine schedule mitigates this effect and distributes the structural information more smoothly along the time steps. Compare Figure~\ref{fig:schedule:alpha}-\ref{fig:schedule:alphabar} for the schedules and Figures~\ref{fig:schedule:linear} and~\ref{fig:schedule:cosine} for a visualization of the respective noising processes. We used the cosine diffusion schedule in all our experiments.

\subsubsection{Different Sampling Methods}\label{subsec:sampling}

We consider three different sampling procedures. The target is always, given a datapoint and a timestep $x_t$, to predict the state of the data point at the precious timestep: $x_{t-1}$, then use an iterative process to sample $x_0$. The most direct way is to train a neural network to directly predict $x_{t-1}$. This is usually done in a variational manner by training to predict the mean value of $p(x_{t-1} | x_t)$. We call this method {\sl single step} sampling. One can also reparameterize the sampling and train the network to predict the clean input $x_0$ and then use the knowledge about the diffusion process to get $x_{t-1}$. We call this method {\sl whole step} sampling. The most sophisticated and commonly used method also uses the information about the diffusion process and reparametrizes the mean of $p(x_{t-1} | x_t)$ the difference between $x_{t-1}$ and $x_t$. As this learns to predict the added noise, we call this method {\sl noise} sampling.

\begin{figure}[tb]
    \centering
    \includegraphics[width=0.19\textwidth]{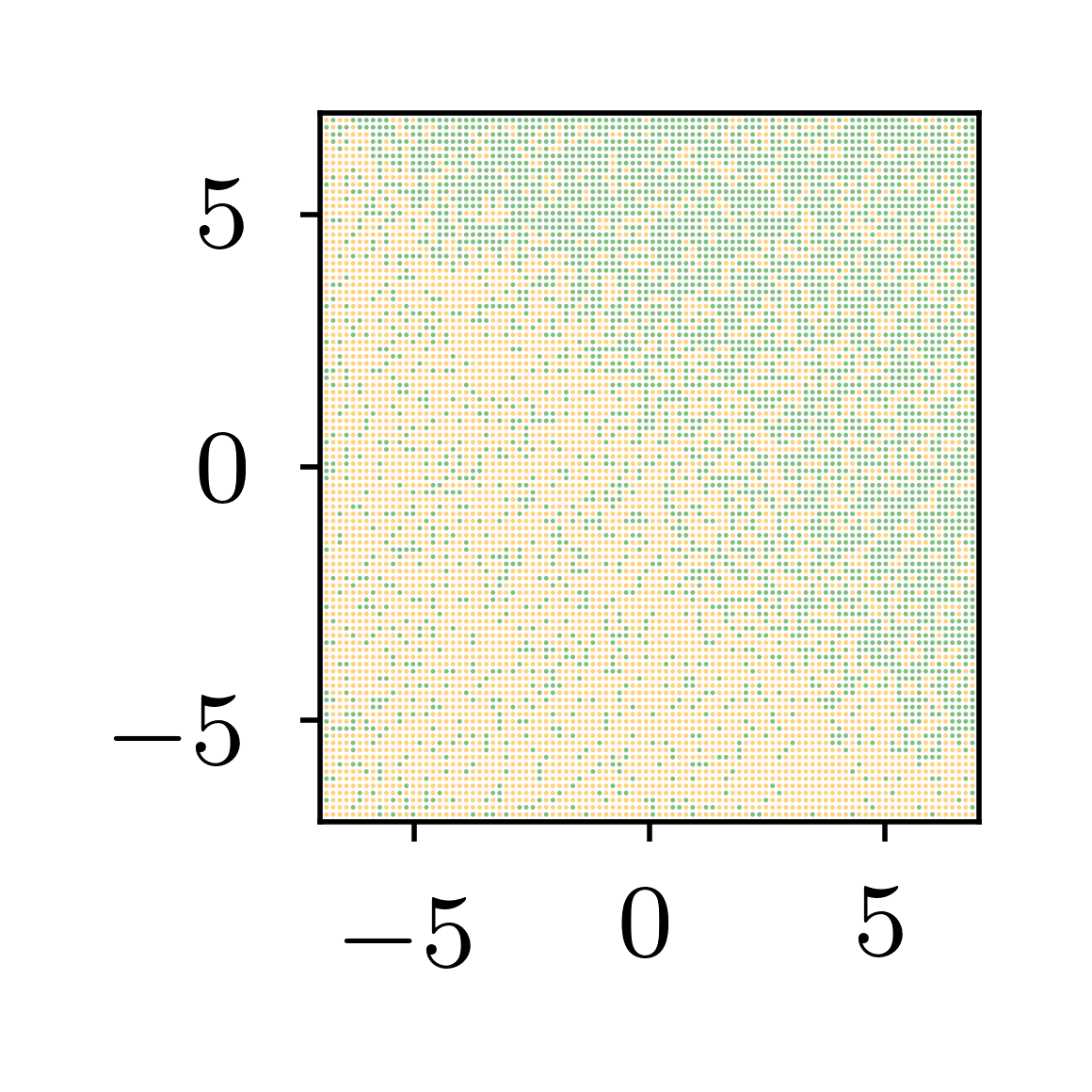}
    \includegraphics[width=0.19\textwidth]{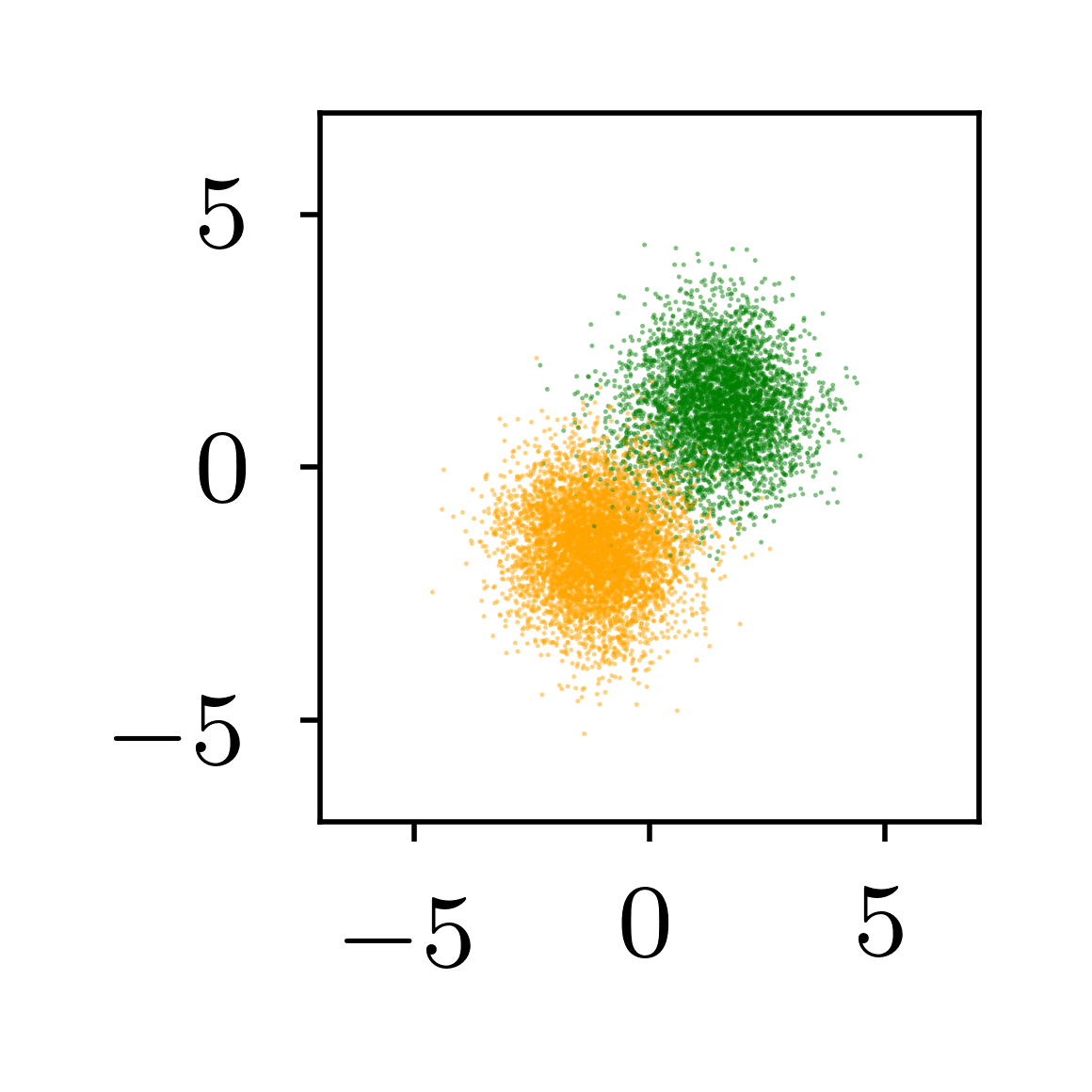}
    \includegraphics[width=0.19\textwidth]{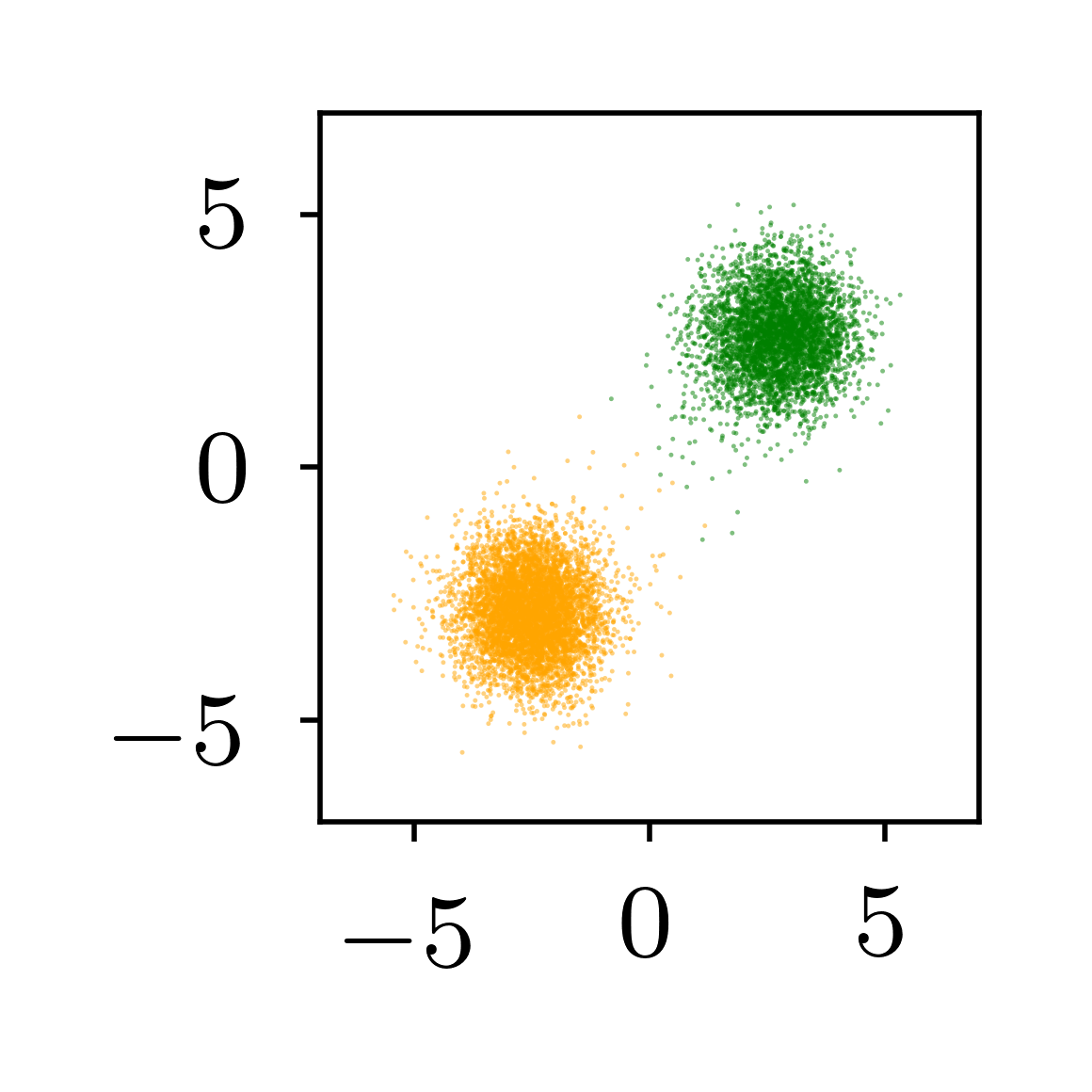}
    \includegraphics[width=0.19\textwidth]{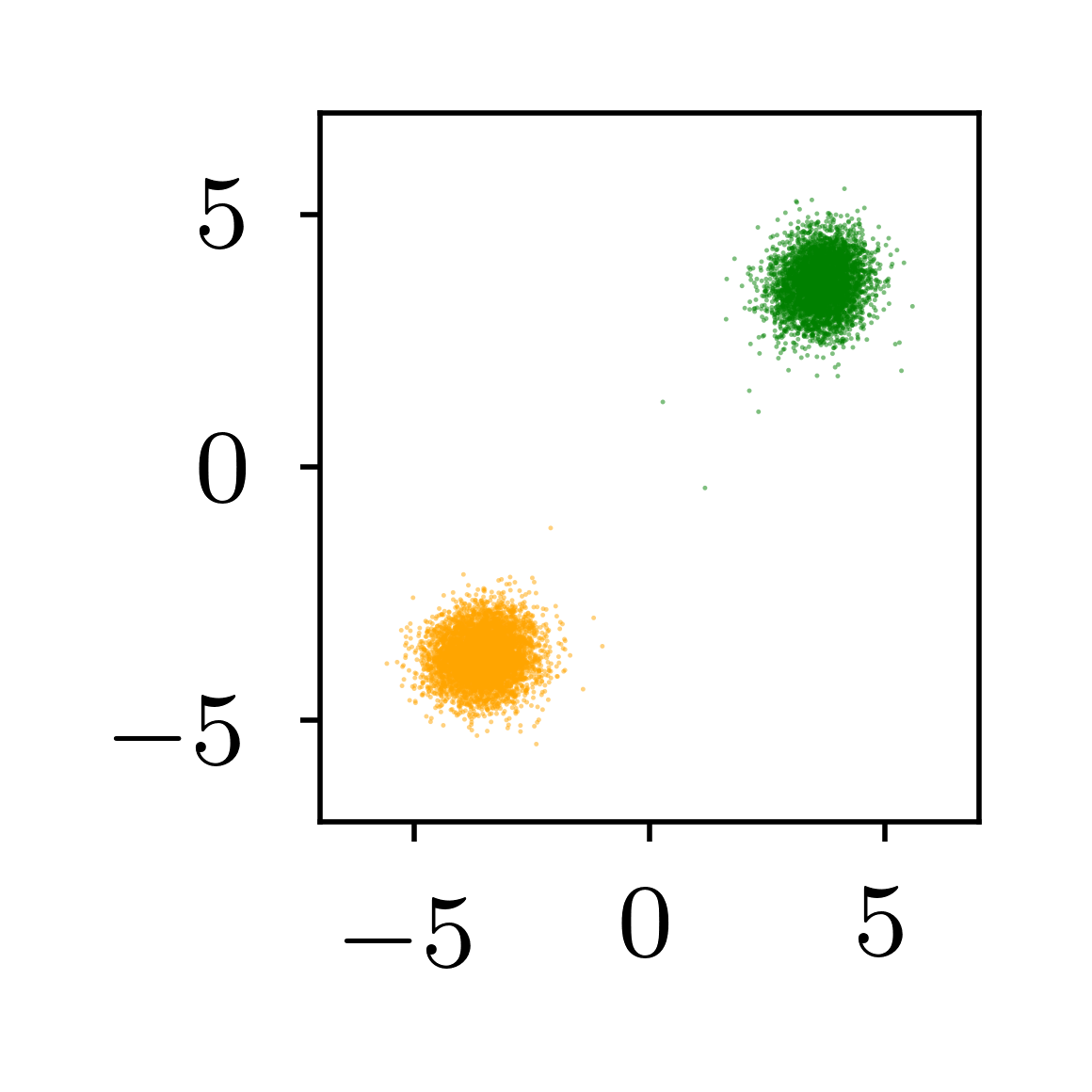}
    \includegraphics[width=0.19\textwidth]{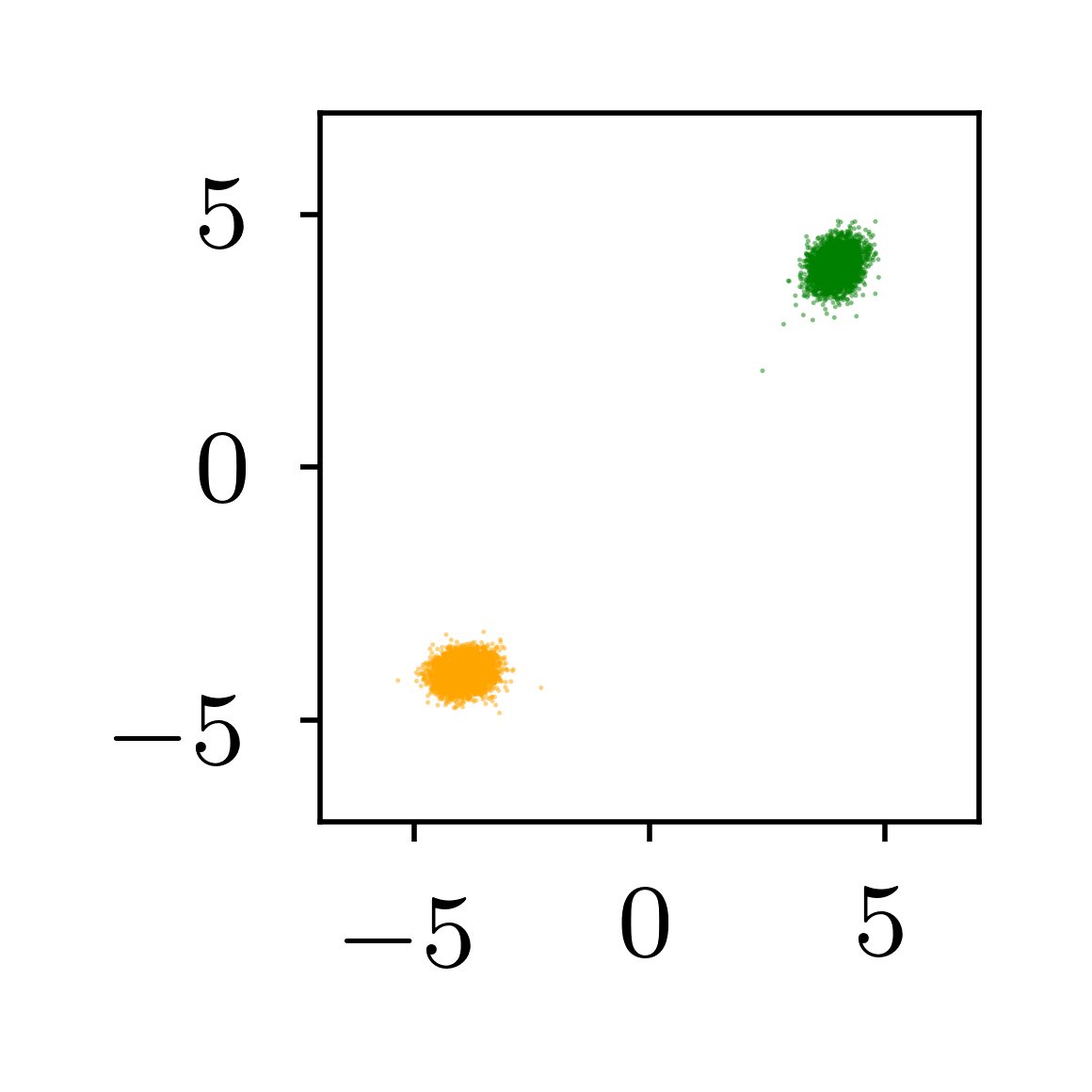}
    \caption{Reparametrized single step denoising process as suggested in \cite{ddpm}. The neural network is trained to approximate the noise in each step.}
    \label{fig:soph:denoise100}
\end{figure}

\begin{figure}[tb]
    \centering
    \includegraphics[width=0.19\textwidth]{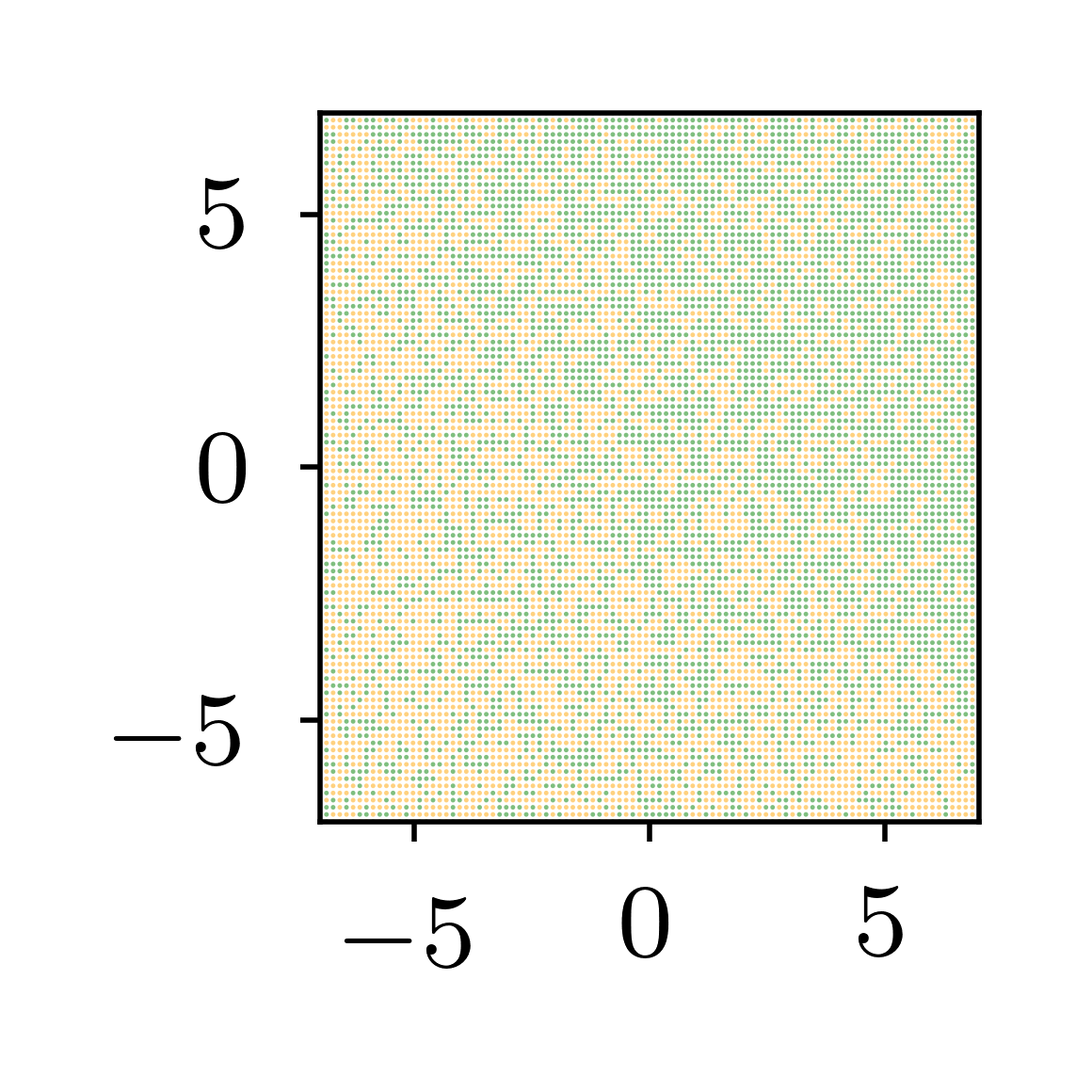}
    \includegraphics[width=0.19\textwidth]{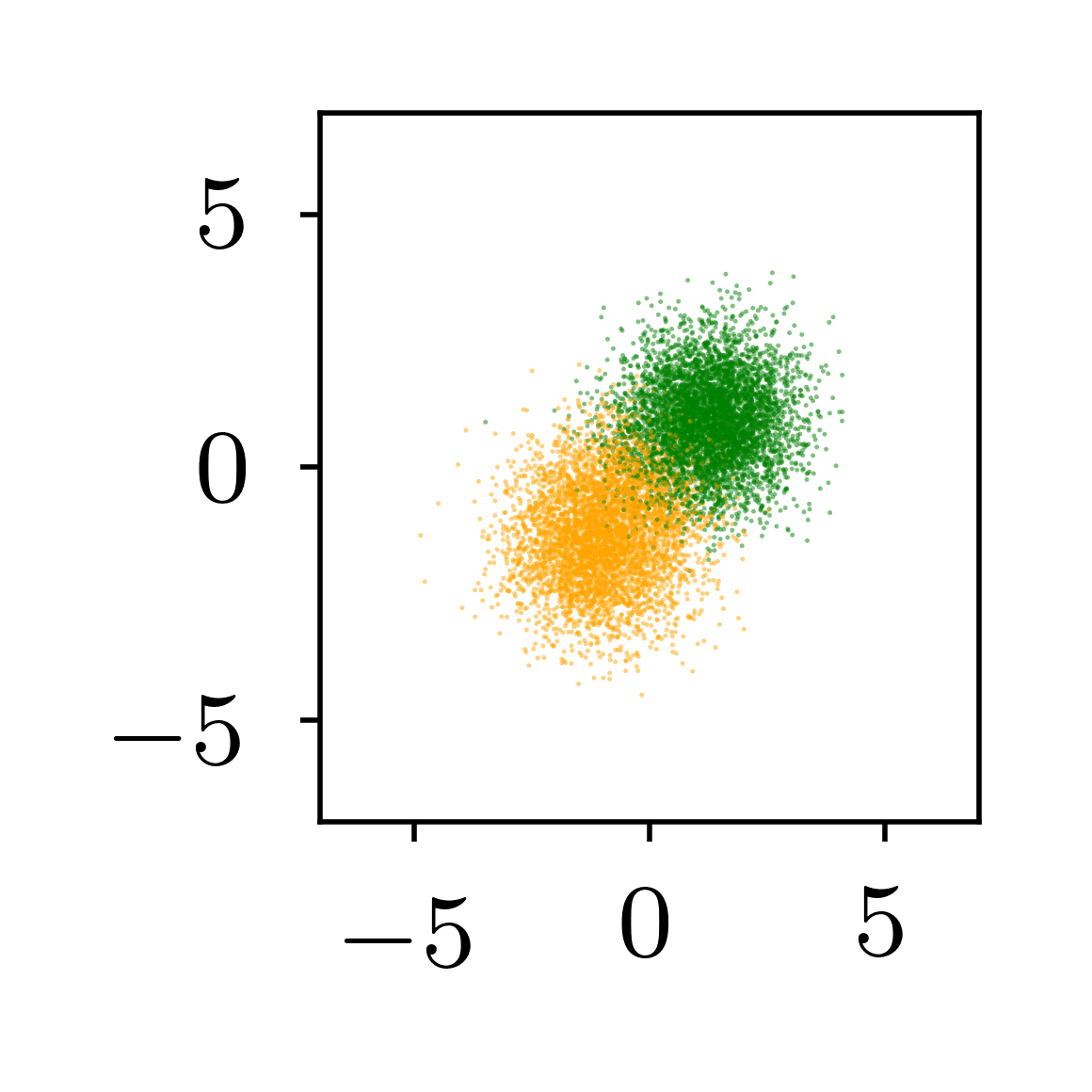}
    \includegraphics[width=0.19\textwidth]{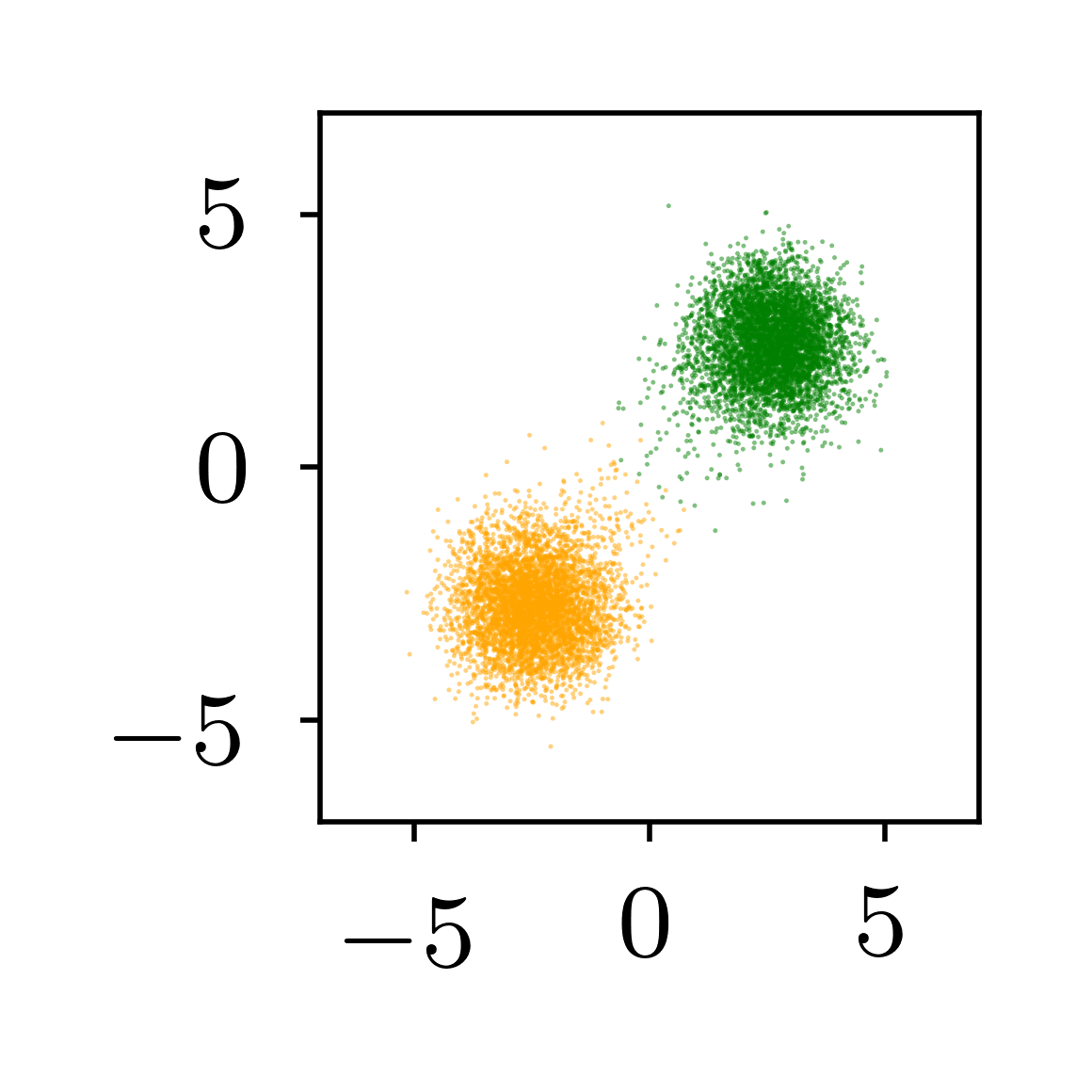}
    \includegraphics[width=0.19\textwidth]{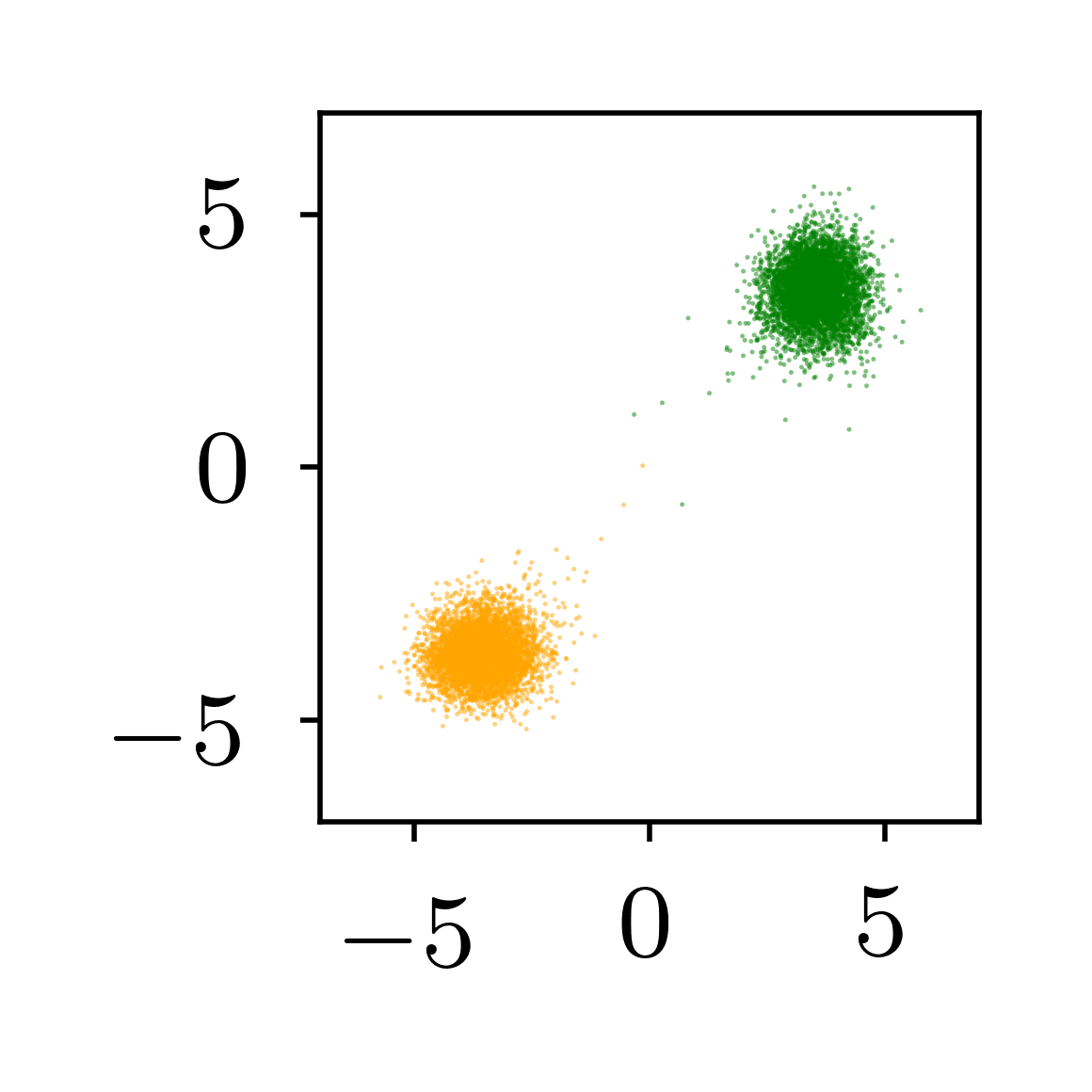}
    \includegraphics[width=0.19\textwidth]{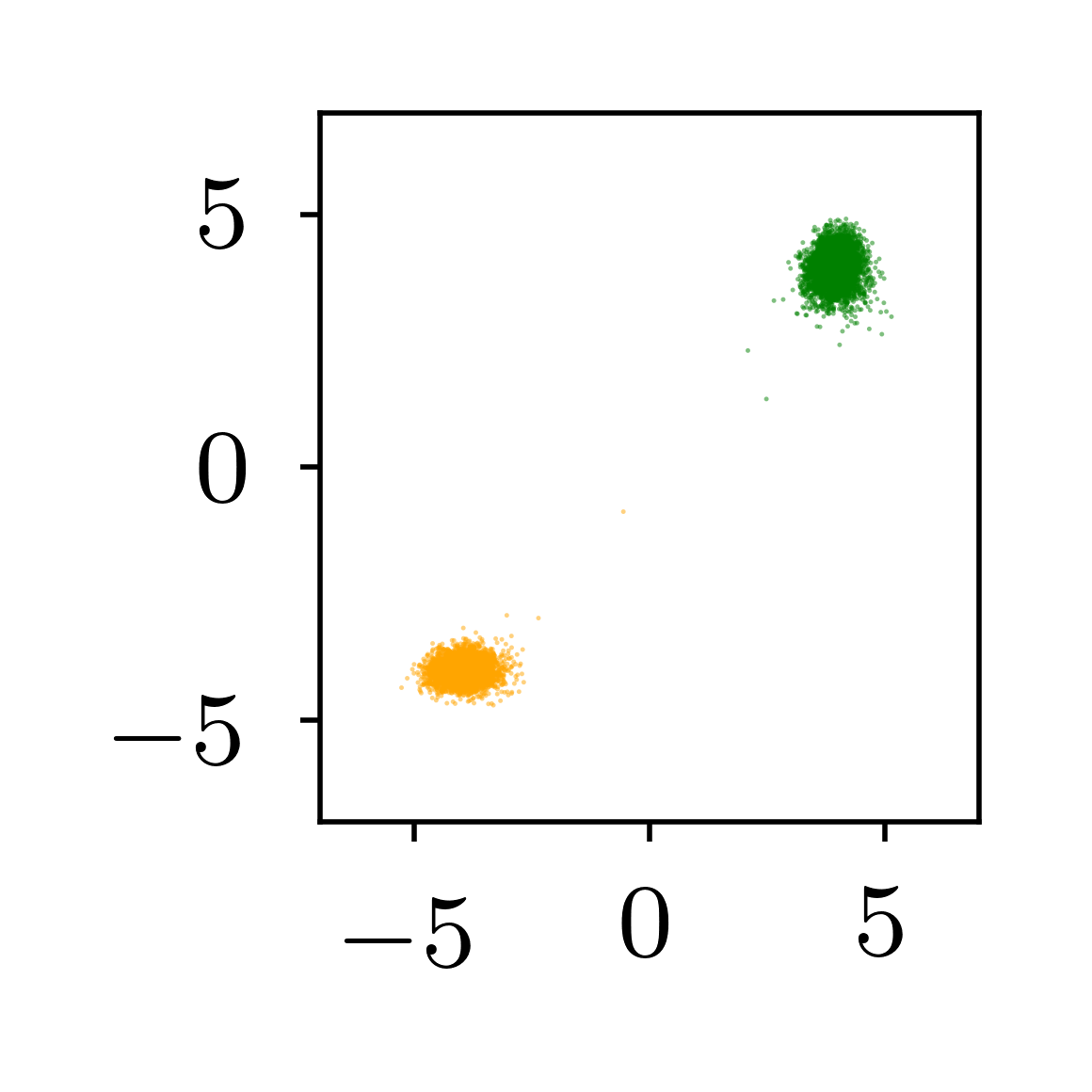}
    \caption{Deterministic noise. Reparametrized single-step denoising process as suggested in \cite{ddpm}. The neural network is trained to approximate the noise in each step.}
    \label{fig:det:denoise100}
\end{figure}

\begin{figure}[tb]
    \centering
    \includegraphics[width=0.19\textwidth]{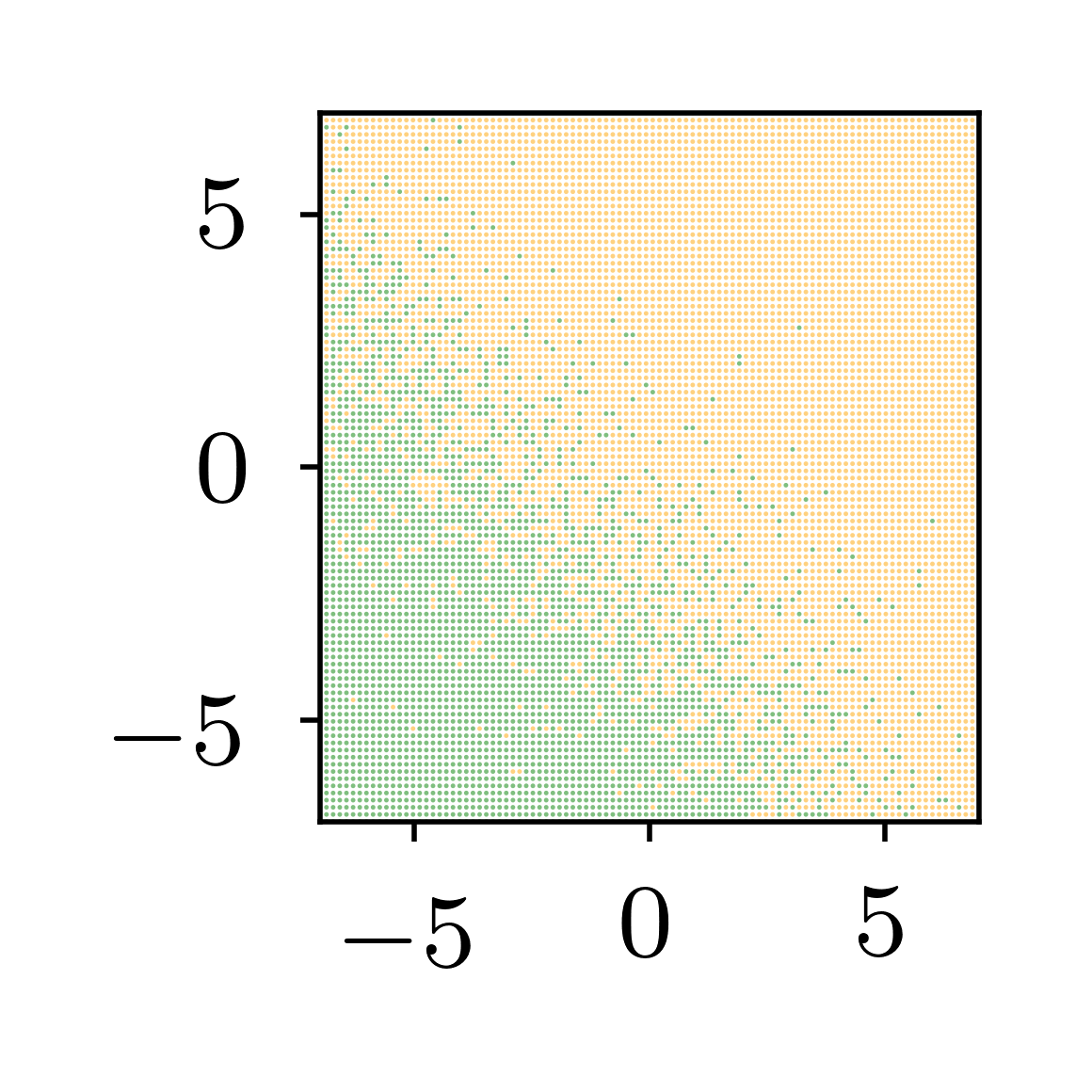}
    \includegraphics[width=0.19\textwidth]{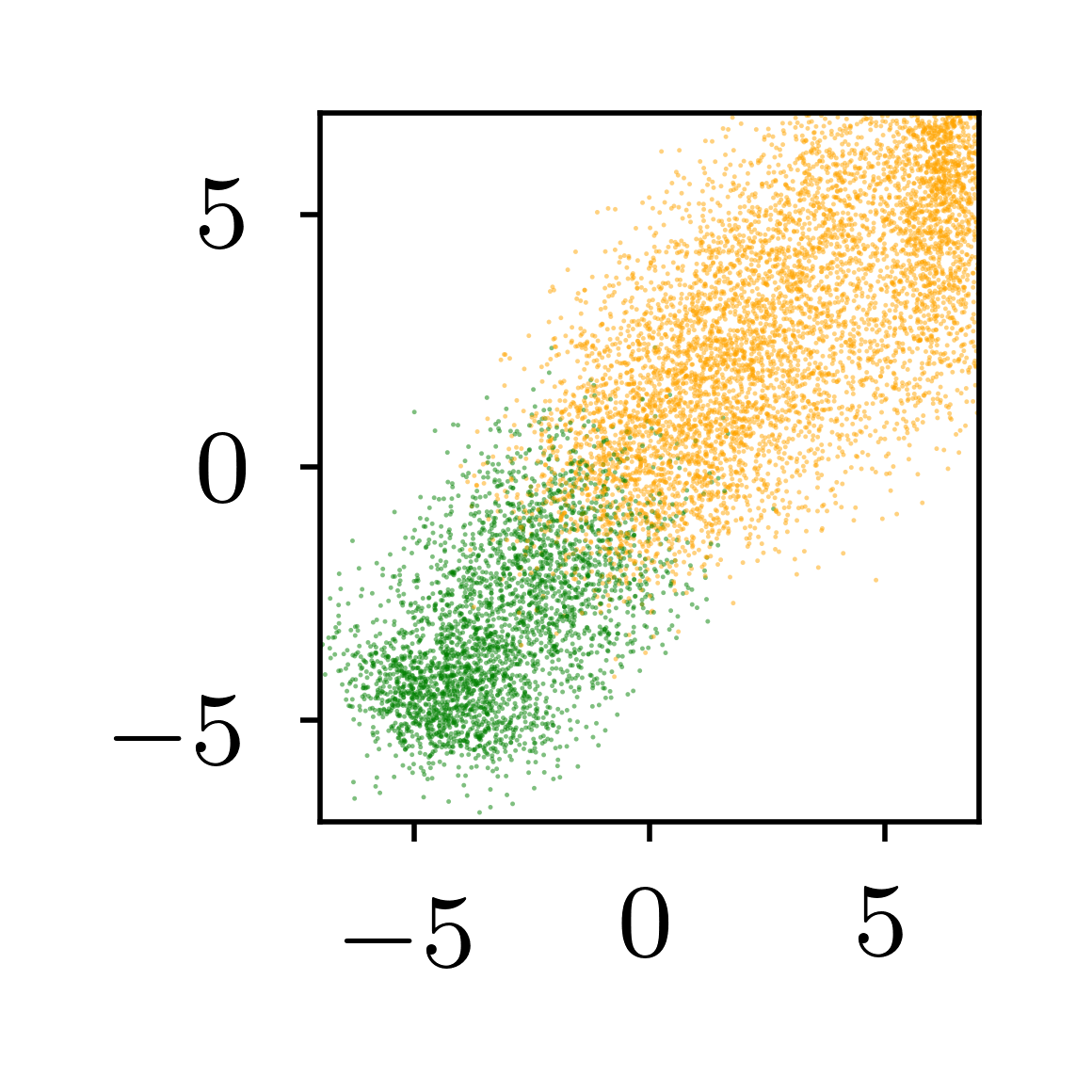}
    \includegraphics[width=0.19\textwidth]{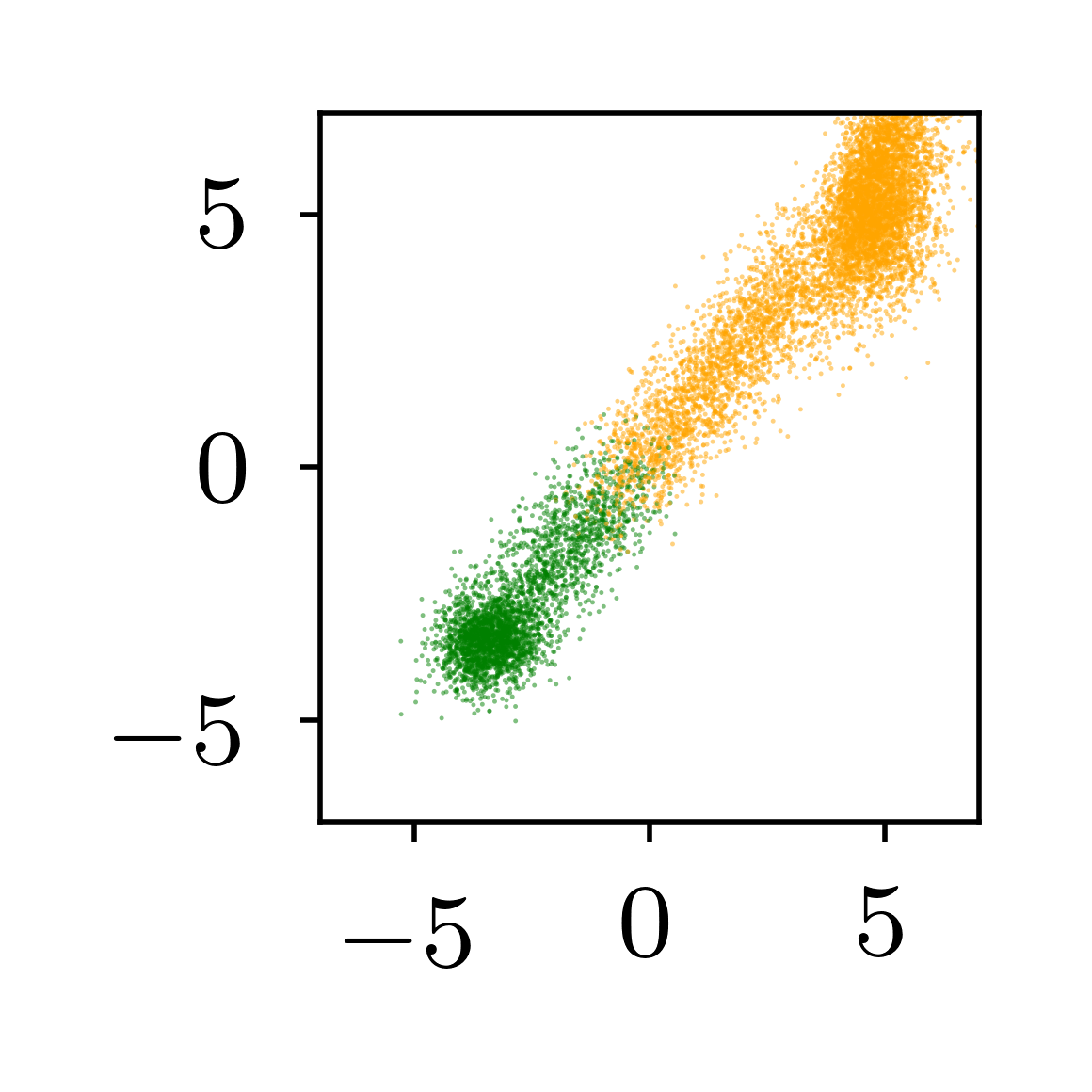}
    \includegraphics[width=0.19\textwidth]{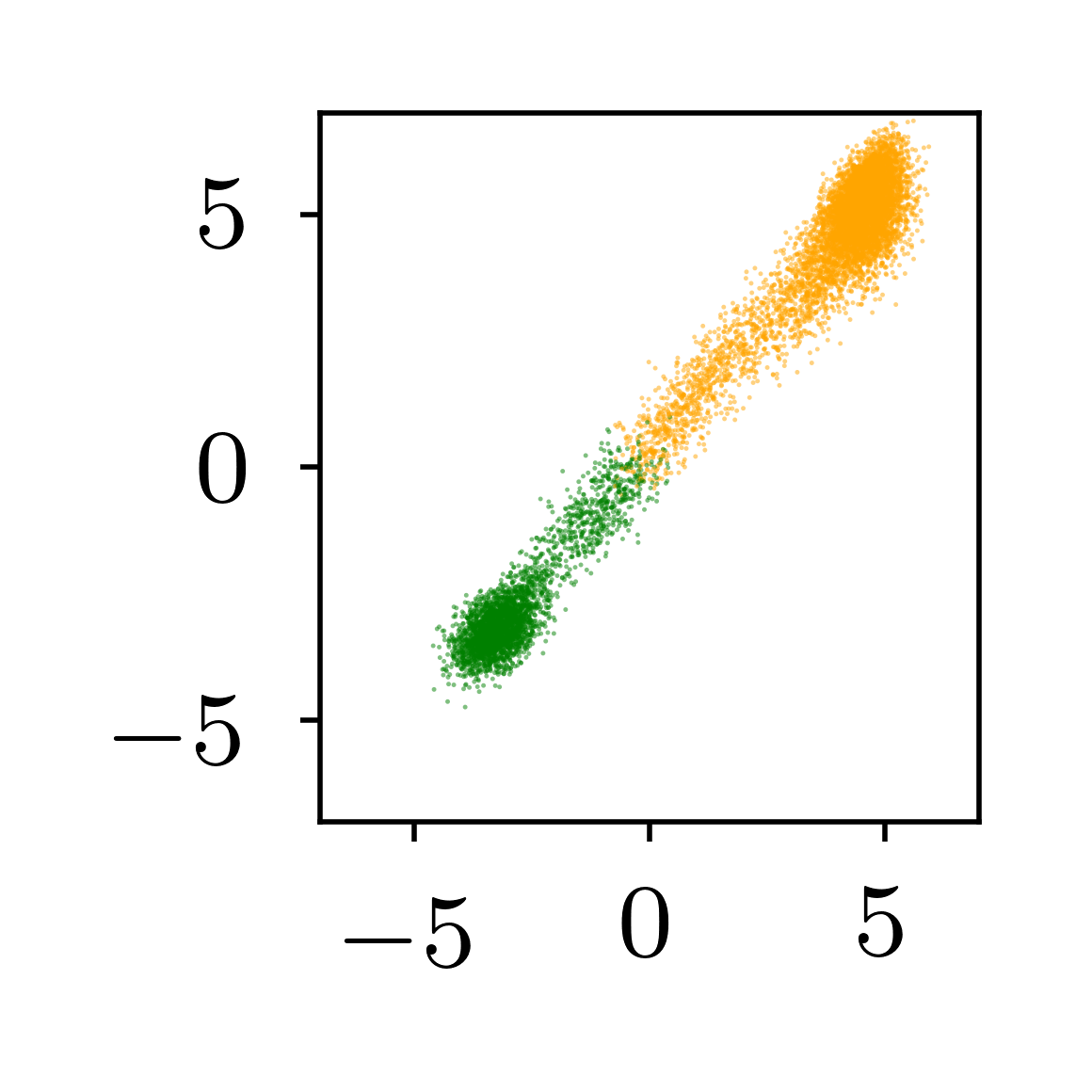}
    \includegraphics[width=0.19\textwidth]{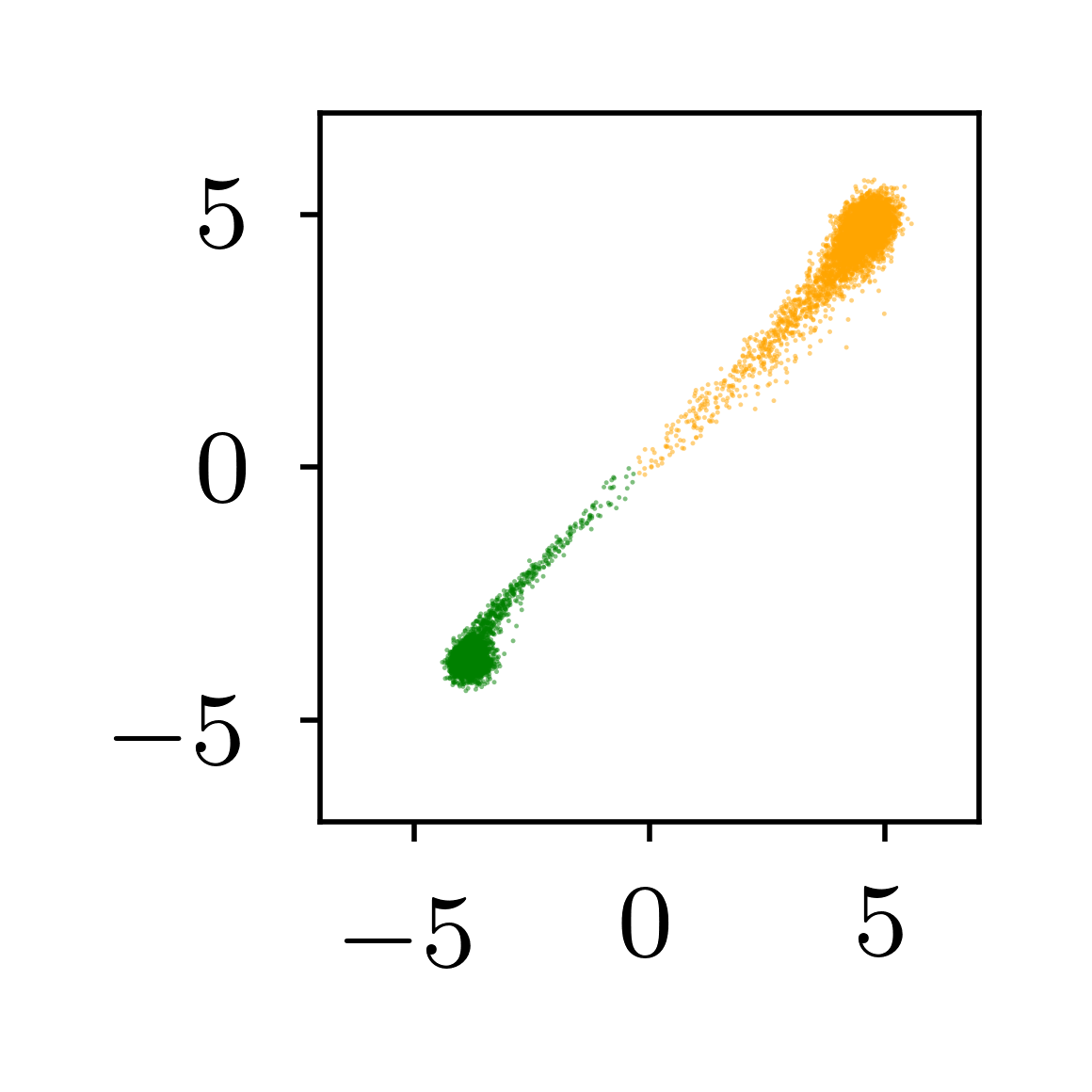}
    \caption{Single step denoising process. The neural network is trained to approximate the mean of the distribution in the previous step $\mu_{t-1}$.}
    \label{fig:single:denoise100}
\end{figure}

\begin{figure}[tb]
    \centering
    \includegraphics[width=0.19\textwidth]{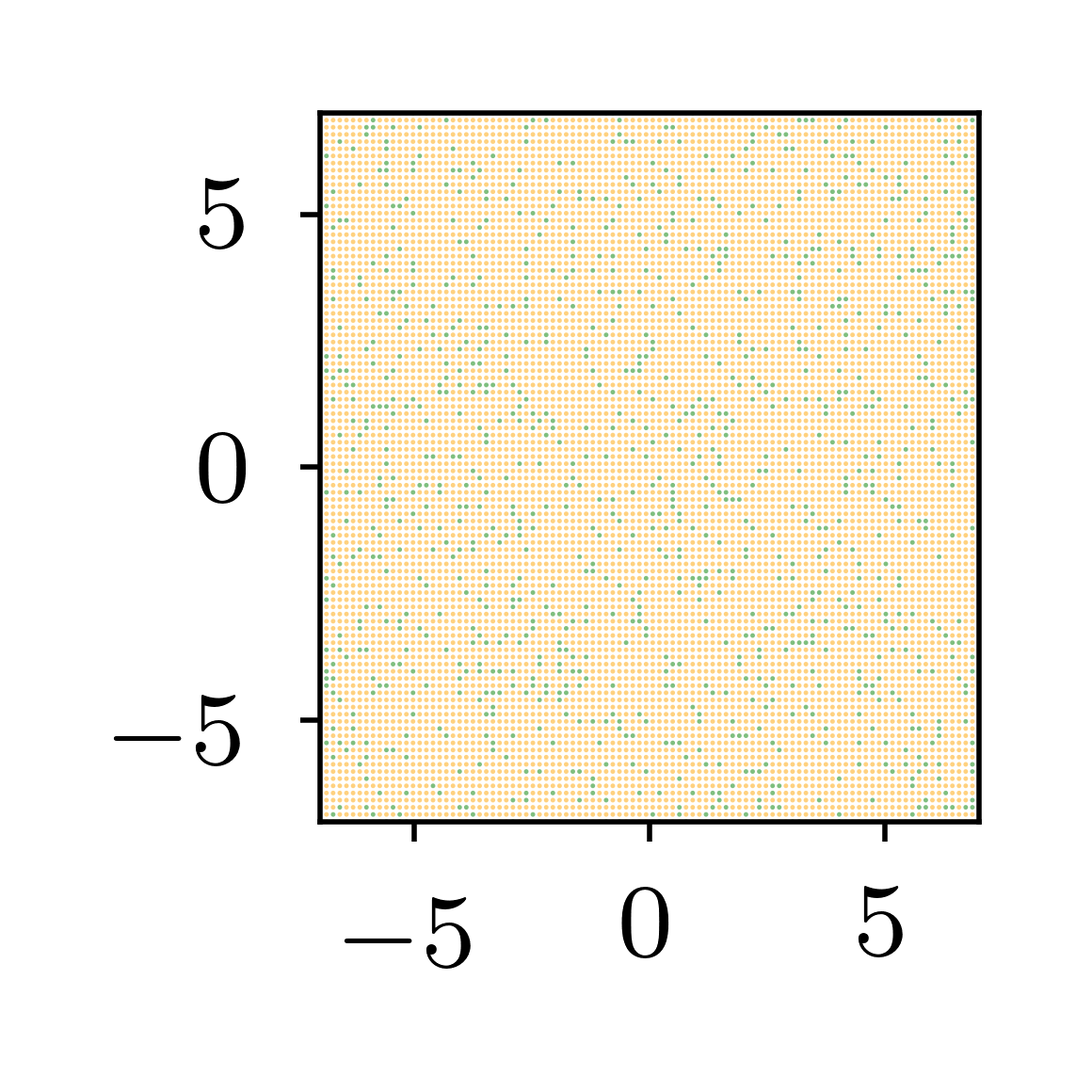}
    \includegraphics[width=0.19\textwidth]{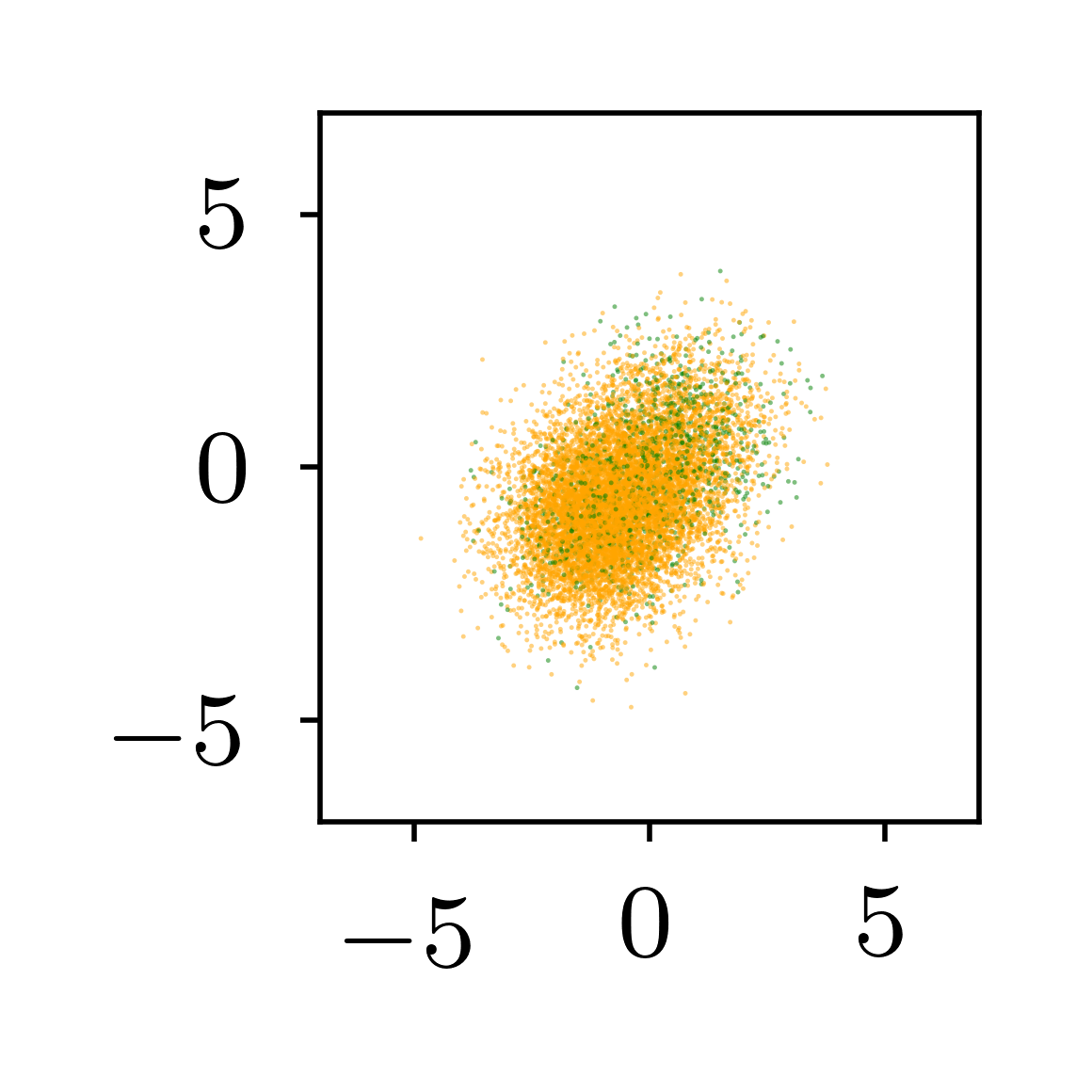}
    \includegraphics[width=0.19\textwidth]{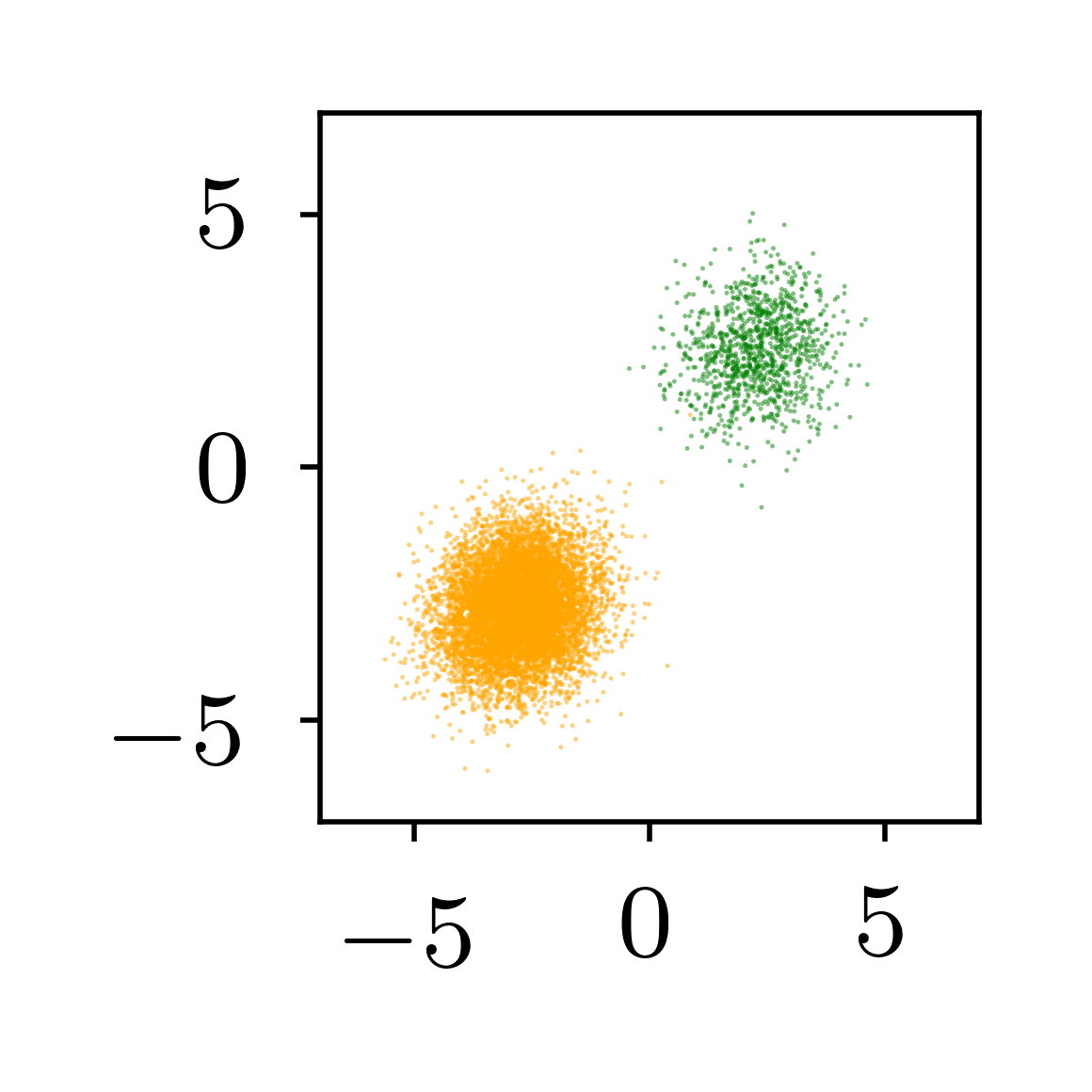}
    \includegraphics[width=0.19\textwidth]{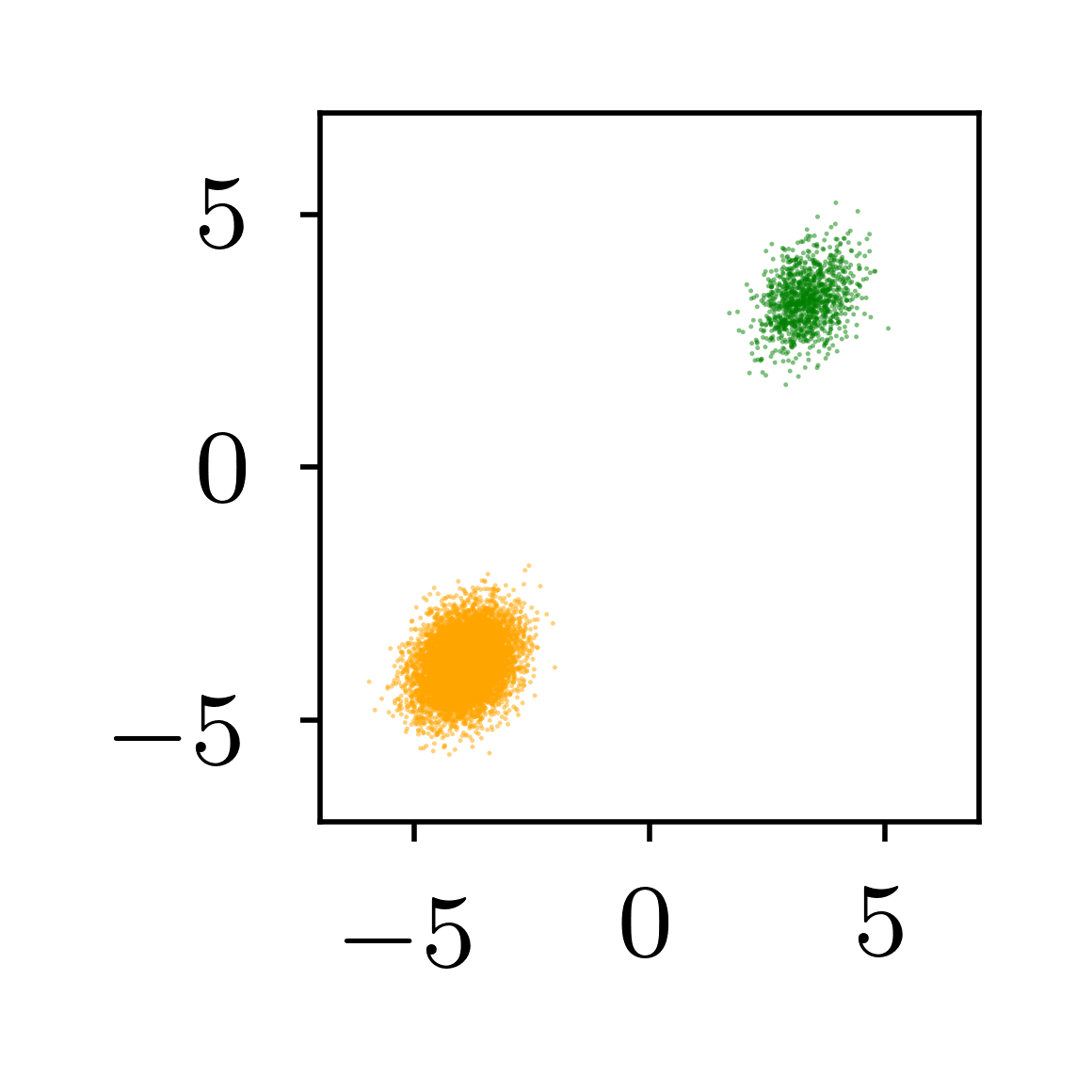}
    \includegraphics[width=0.19\textwidth]{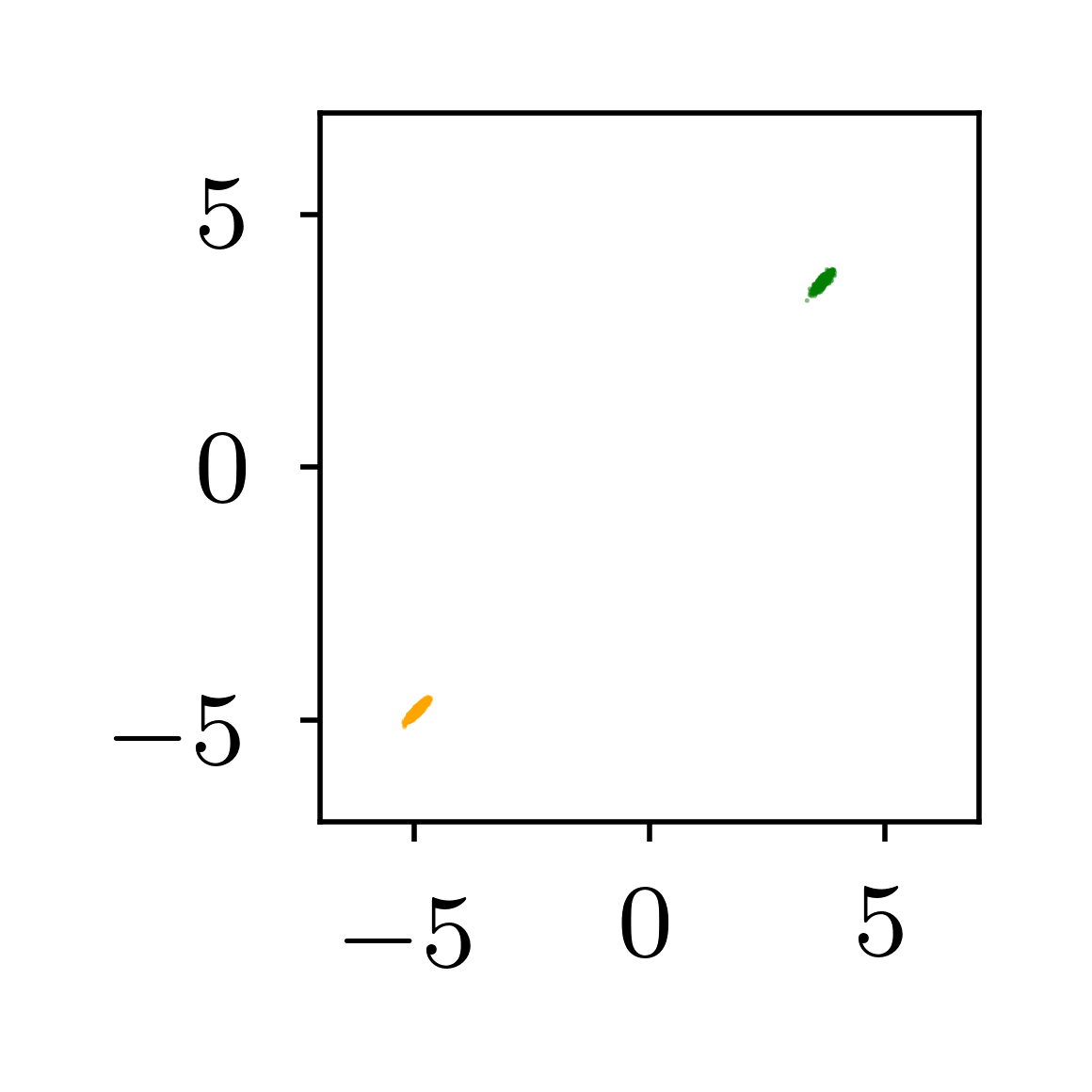}
    \caption{Whole step denoising process. The neural network is trained to approximate the clean image $x_0$. $x_{t-1}$ is sampled by adding $t-1$ steps of noise process.}
    \label{fig:whole:denoise100}
\end{figure}

\begin{figure}[ht]
    \input{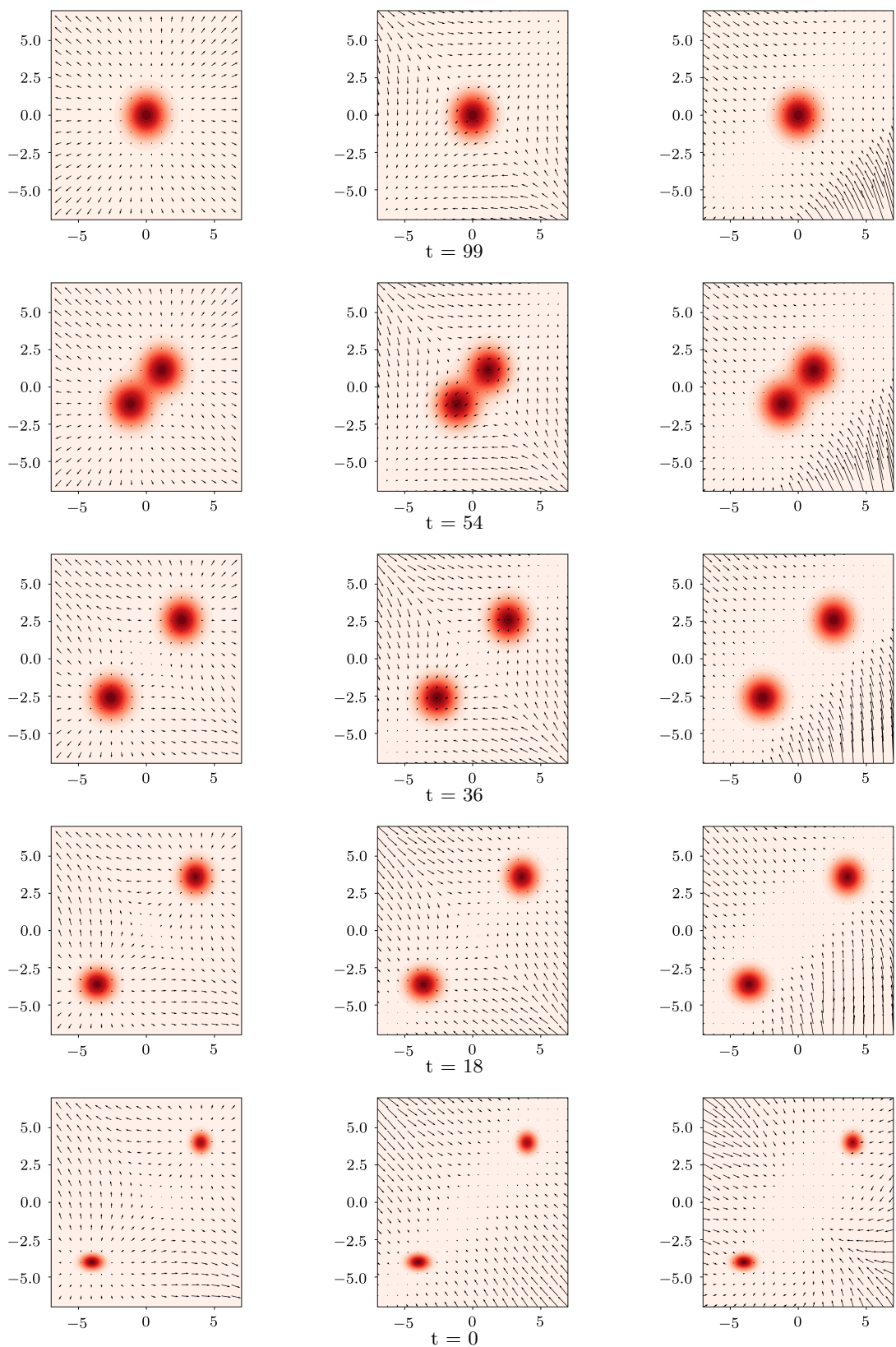}
    
    \caption{The function approximated by the neural network for the three different sampling methods. The arrows indicate the approximated function when trained to predict the noise $\epsilon_t$ (left), when trained to predict the clean image $x_0$ (middle), and when trained to predict the previous time step $x_{t-1}$ (right).}
    \label{fig:fun:100}
\end{figure}

\begin{figure}[ht]
    \input{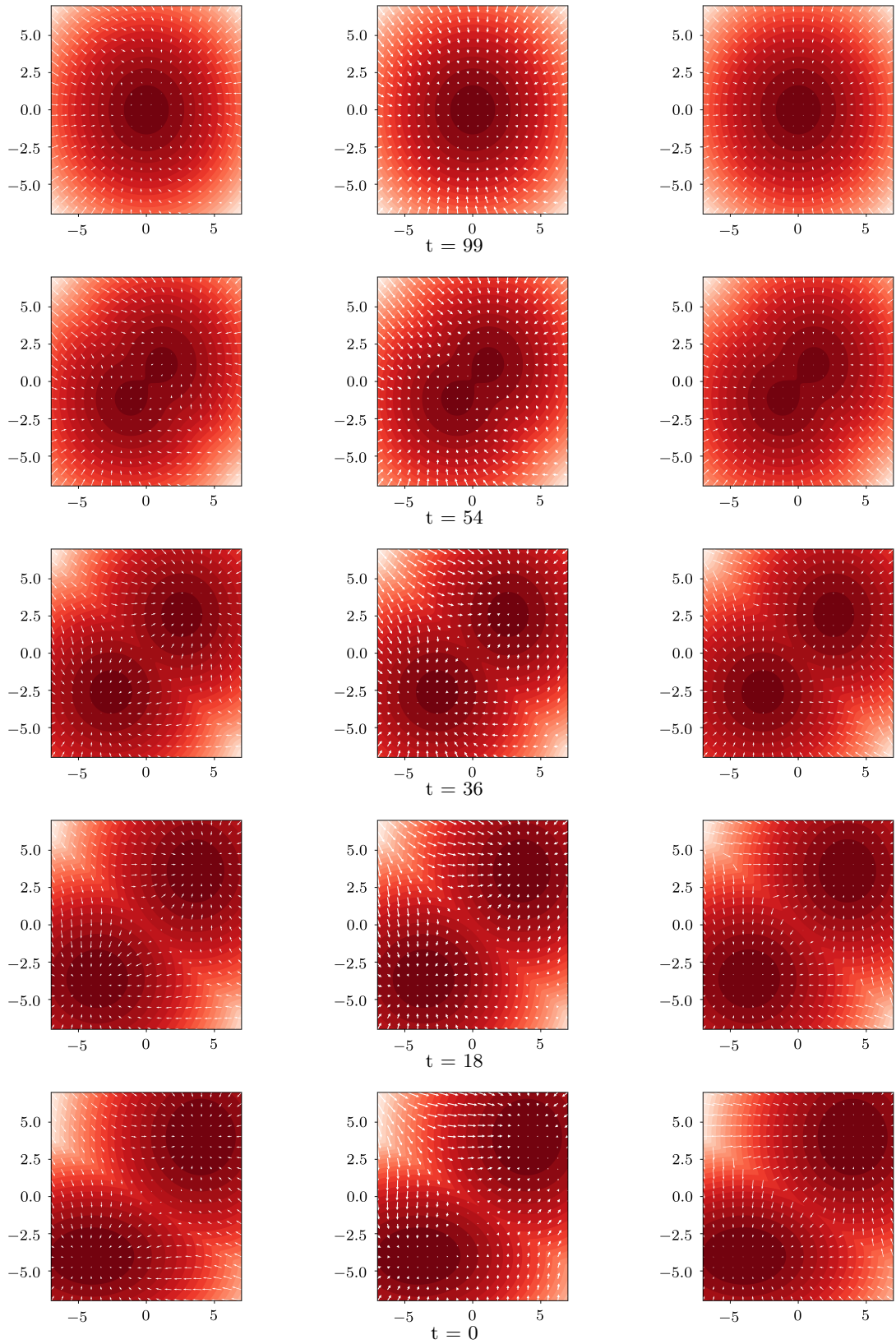}
    
    \caption{Comparison of the learned trajectories for the random noise versus the noiseless diffusion process. The arrows indicate the final trajectories for the standard model trained on the diffusion process using random noise (left), for a model trained on a deterministic diffusion process (middle), and for comparison, the true score of the function (right). The score is proportional to the noise given a fixed point.}
    \label{fig:traj:100}
\end{figure}

\subsubsection{Does the Process invert the diffusion?}

We visualize the denoising processes for the different sampling methods in Figures~\ref{fig:soph:denoise100} to~\ref{fig:whole:denoise100}. The figures show time steps t = 99, 54, 36, 18, and 0 for 10000 data points. The data points for time step 99 are taken from a grid between -7 and 7. The colors in the figures indicate group membership when clustering with a Gaussian Mixture Model on the final time step.

While the {\sl noise} sampling approximates the ground truth distribution reasonably well, the other sampling methods fail to fit the training distribution. We observe many samples from the low-density region between the two clusters for the {\sl single step} sampling. The {\sl whole step} sampling only samples points from very high-density regions and almost collapses to the means of the two Gaussians.

Most surprising is the significantly worse performance of the {\sl single step} sampling compared to the {\sl noise} sampling, as this is a simple reparametrization. However, we explicitly add information about the forward diffusion process when sampling the noise relative to the clean data instead of the data point at the previous timestep. This information is hard to infer for the model trained on this data with only implicit access to this information.

Another phenomenon we observe is that the two first sampling methods keep a positional bias, so points closer to $ \mu_1 $ in the beginning end up close to $ \mu_1 $ in the end. The whole-step sampling methods suffer less from this phenomenon. Also, the model trained with the deterministic diffusion process shows no positional bias. This shows that even though, for the {\sl noise} sampling, at first glance, one might think the reverse process is approximated, it is not. This is not possible for random noise, so the intuition of reversing the diffusion is misleading. However, this also indicates that the model, to some degree, fits the data gradient. In the following, we do further simulations to investigate what precisely the models learn.

\subsubsection{What does the neural network approximate?}

Figure~\ref{fig:fun:100} provides insights into the neural network's learning outcomes for the three specific objectives discussed in Section~\ref{subsec:sampling}. The network struggles to accurately approximate the score function across various regions in all tasks. This is expected as the model deals with very noisy data in the first time steps and sparse regions in the latter. Especially in the low-density regions in the last time step, we can not expect the model to learn a helpful function as training data in those areas is limited. The necessity of the iterative sampling process becomes evident in these images, showcasing that only when combining and aggregating the information available at individual time steps results in a sufficient approximation of the training distribution. 

This observation underscores the significant effect of introducing stepwise perturbations into the training process to ensure adequate coverage of low-density regions and the effective learning of data distribution gradients. \citet{song2019genmodelscore} also observe this behavior and argue that noise is the solution for a good approximation of the score function. In the following experiment, we showcase that this is only one perspective.

\subsubsection{What about Noise?}

If the coverage of low-density regions is the core problem, then the diffusion process, not noise itself, is the key to a good approximation of the score function. We define a deterministic diffusion process that modifies a data point in each step to converge to a normal distribution. For every data point, we take the x'ths number after the comma. These values are approximately uniformly distributed. To go from the uniform distribution to a normal distribution, we map it through the inverse of the cdf. The resulting diffusion process when using the cosine schedule is shown in Figure~\ref{fig:det:denoise100}. 

We train the same model with the same hyperparameters as before and observe similar behavior and performance on the three sampling methods. Figure~\ref{fig:det:denoise100} shows the learned trajectories for the reparametrized denoising process. 

We conclude that diffusion is necessary to cover the whole space. However, we can do this in an unnoisy way. If we could construct a deterministic diffusion process that is also invertible, we could achieve perfect recovery of training data while still being able to sample new data points.

While \cite{bansal2024cold} also observes good performance for deterministic diffusion processes, they consider a very different sampling setting, and thus, their insights do not translate to our setting; they do not aim to approximate the score gradient of data but the datapoint at the previous timestep. So, their target is not the approximate score. This would be difficult in low-density regions and would not work using their ``diffusion'' processes.

\newpage
\section{What's next}

In considering the future directions for our research, several intriguing questions emerge, separating into two overarching areas.

\subsection{Noise vs. No Noise}

If we decide we do not need any noise in the diffusion pipeline, what are the benefits and drawbacks of using it? An essential consideration is the computational efficiency of computing the diffusion deterministically. This is computationally more expensive in our current approach and prompts evaluating whether the computational overhead is justified and what advantages deterministic diffusion may bring.

Additionally, exploring the implications of training on deterministic data remains an exciting question. The prospect of achieving perfect recovery of training data through deterministic training raises the question: is this even desired as the goal of these models is to generate new data? What are the limitations or (unwanted) biases introduced to the model compared to random noise?

\subsection{Take it back to graphs}

Shifting our focus back to the domain of graphs introduces a distinctive set of challenges and considerations. Unlike images, which are essentially high-dimensional vectors, graphs encapsulate diverse and heterogeneous forms of information. Notably, distributions over molecules present challenges regarding description and analysis. The central question at hand involves understanding what the "score function" captures in graph data and critically assessing how well we are approximating the underlying distribution.

Delving deeper, a key question is understanding what information our model learns about graphs. Is the structural information captured sufficiently even though it is not explicitly included in the diffusion process? Determining what the model learns well and what it might miss is crucial, especially considering how complex graph data can be.

Moreover, the unique way the graphs sampling is introduced in \cite{digress} calls for further exploration. Figuring out why predicting a clear graph works better than predicting a clear image could help us improve our understanding of the model and the model itself, especially when dealing with different data types.

\newpage
\bibliography{main}

\begin{thebibliography}{11}
\providecommand{\natexlab}[1]{#1}
\providecommand{\url}[1]{\texttt{#1}}
\expandafter\ifx\csname urlstyle\endcsname\relax
  \providecommand{\doi}[1]{doi: #1}\else
  \providecommand{\doi}{doi: \begingroup \urlstyle{rm}\Url}\fi

\bibitem[Austin et~al.(2021)Austin, Johnson, Ho, Tarlow, and Van Den~Berg]{austin2021structured}
J.~Austin, D.~D. Johnson, J.~Ho, D.~Tarlow, and R.~Van Den~Berg.
\newblock Structured denoising diffusion models in discrete state-spaces.
\newblock 2021.

\bibitem[Bansal et~al.(2024)Bansal, Borgnia, Chu, Li, Kazemi, Huang, Goldblum, Geiping, and Goldstein]{bansal2024cold}
A.~Bansal, E.~Borgnia, H.-M. Chu, J.~Li, H.~Kazemi, F.~Huang, M.~Goldblum, J.~Geiping, and T.~Goldstein.
\newblock Cold diffusion: Inverting arbitrary image transforms without noise.
\newblock Neural Information Processing Systems (NeurIPS), 2024.

\bibitem[Haefeli et~al.(2022)Haefeli, Martinkus, Perraudin, and Wattenhofer]{graphsdiscrete2022diffusion}
K.~K. Haefeli, K.~Martinkus, N.~Perraudin, and R.~Wattenhofer.
\newblock Diffusion models for graphs benefit from discrete state spaces.
\newblock NeurIPS Workshop on New Frontiers in Graph Learning, 2022.

\bibitem[Ho et~al.(2020)Ho, Jain, and Abbeel]{ddpm}
J.~Ho, A.~Jain, and P.~Abbeel.
\newblock Denoising diffusion probabilistic models.
\newblock Neural Information Processing Systems (NeurIPS), 2020.

\bibitem[Lim et~al.(2023)Lim, Kovachki, Baptista, Beckham, Azizzadenesheli, Kossaifi, Voleti, Song, Kreis, Kautz, et~al.]{lim2023function}
J.~H. Lim, N.~B. Kovachki, R.~Baptista, C.~Beckham, K.~Azizzadenesheli, J.~Kossaifi, V.~Voleti, J.~Song, K.~Kreis, J.~Kautz, et~al.
\newblock Score-based diffusion models in function space.
\newblock \emph{arXiv preprint arXiv:2302.07400}, 2023.

\bibitem[Montanari(2023)]{stochstic_loc}
A.~Montanari.
\newblock Sampling, diffusions, and stochastic localization.
\newblock \emph{arXiv preprint arXiv:2305.10690}, 2023.

\bibitem[Nichol and Dhariwal(2021)]{ddpm2021improved}
A.~Q. Nichol and P.~Dhariwal.
\newblock Improved denoising diffusion probabilistic models.
\newblock International Conference on Machine Learning, 2021.

\bibitem[Sohl-Dickstein et~al.(2015)Sohl-Dickstein, Weiss, Maheswaranathan, and Ganguli]{thermodynamics}
J.~Sohl-Dickstein, E.~Weiss, N.~Maheswaranathan, and S.~Ganguli.
\newblock Deep unsupervised learning using nonequilibrium thermodynamics.
\newblock International Conference on Machine Learning (ICML), 2015.

\bibitem[Song and Ermon(2019)]{song2019genmodelscore}
Y.~Song and S.~Ermon.
\newblock Generative modeling by estimating gradients of the data distribution.
\newblock Neural Information Processing Systems (NeurIPS), 2019.

\bibitem[Song et~al.(2021)Song, Sohl-Dickstein, Kingma, Kumar, Ermon, and Poole]{song2021scorebased}
Y.~Song, J.~Sohl-Dickstein, D.~P. Kingma, A.~Kumar, S.~Ermon, and B.~Poole.
\newblock Score-based generative modeling through stochastic differential equations.
\newblock International Conference on Learning Representations (ICLR), 2021.

\bibitem[Vignac et~al.(2023)Vignac, Krawczuk, Siraudin, Wang, Cevher, and Frossard]{digress}
C.~Vignac, I.~Krawczuk, A.~Siraudin, B.~Wang, V.~Cevher, and P.~Frossard.
\newblock Digress: Discrete denoising diffusion for graph generation.
\newblock International Conference on Learning Representations, 2023.

\end{thebibliography}

\end{document}